\documentclass{article}

\usepackage{microtype}
\usepackage{graphicx}
\usepackage{booktabs} 

\usepackage{tabularx}
\usepackage{array}
\usepackage{makecell}

\usepackage{tcolorbox}
\tcbuselibrary{listingsutf8}

\usepackage{url}

\usepackage{color}
\usepackage{listings} 
\usepackage[frozencache,cachedir=minted-cache]{minted} 

\usepackage{mdframed}
\usepackage{array}

\usepackage{eqparbox}
\usepackage{caption}
\usepackage{subcaption}
\usepackage{tikz}
\usepackage{amsmath}
\usepackage{amsthm,thmtools}
\usepackage{multirow}
\usepackage{tabularx}

\declaretheoremstyle[
  headfont=\bfseries,
  bodyfont=\itshape,
]{myplain}
\declaretheoremstyle[
  headfont=\bfseries,
  bodyfont=\normalfont,
]{mydefinition}
\declaretheoremstyle[
  headfont=\itshape,
  bodyfont=\normalfont,
]{myremark}

\setminted[sql]{breaklines, framesep=3mm, fontsize=\footnotesize, numbersep=5pt}

\definecolor{codegreen}{rgb}{0,0.6,0}
\definecolor{codegray}{rgb}{0.5,0.5,0.5} 
\definecolor{codepurple}{rgb}{0.58,0,0.82}
\definecolor{backcolour}{rgb}{0.95,0.95,0.92}

\lstdefinestyle{mystyle}{
    backgroundcolor=\color{backcolour},   
    commentstyle=\color{codegreen},
    keywordstyle=\color{magenta},
    numberstyle=\tiny\color{codegray},
    stringstyle=\color{codepurple},
    basicstyle=\ttfamily\footnotesize,
    breakatwhitespace=false,         
    breaklines=true,                 
    captionpos=b,                    
    keepspaces=true,                 
    numbers=left,                    
    numbersep=5pt,                  
    showspaces=false,                
    showstringspaces=false,
    showtabs=false,                  
    tabsize=2,
    lineskip=-1pt
}

\newif\ifcomments
\ifcomments
    \providecommand{\ion}[1]{{\color{teal}{[ion: #1]}}}
    
    \providecommand{\joey}[1]{{\color{magenta}{[joey: #1]}}}
    \providecommand{\shu}[1]{{\color{cyan}{[shu: #1]}}}
    
    \providecommand{\asim}[1]{{\color{blue}{[asim: #1]}}}
    \providecommand{\simon}[1]{{\color{orange}{[simon: #1]}}}
    \providecommand{\accheng}[1]{{\color{purple}{[accheng: #1]}}}

\else
    \providecommand{\ion}[1]{}
    \providecommand{\joey}[1]{}
    \providecommand{\shu}[1]{}
    \providecommand{\asim}[1]{}
    \providecommand{\simon}[1]{}
    \providecommand{\accheng}[1]{}
\fi
\newcommand{\SYS}{}
\def\SYS/{SYS}

\usepackage{xspace} 
\newcommand{\optimal}{OPHR\xspace}
\newcommand{\greedy}{GGR\xspace}

\usepackage{hyperref}

\usepackage[accepted]{mlsys2025}

\begin{document}
\lstset{style=mystyle}


\twocolumn[
\mlsystitle{Optimizing LLM Queries in Relational Data Analytics Workloads}



\mlsyssetsymbol{equal}{*}

\begin{mlsysauthorlist}
\mlsysauthor{Shu Liu}{equal,to}
\mlsysauthor{Asim Biswal}{equal,to}
\mlsysauthor{Amog Kamsetty}{to}
\mlsysauthor{Audrey Cheng}{to}
\mlsysauthor{Luis Gaspar Schroeder}{to,goo}
\end{mlsysauthorlist}
\begin{mlsysauthorlist}
\mlsysauthor{Liana Patel}{ed}
\mlsysauthor{Shiyi Cao}{to}
\mlsysauthor{Xiangxi Mo}{to}
\mlsysauthor{Ion Stoica}{to}
\mlsysauthor{Joseph E. Gonzalez}{to}
\mlsysauthor{Matei Zaharia}{to}
\end{mlsysauthorlist}

\mlsysaffiliation{to}{UC Berkeley}
\mlsysaffiliation{goo}{Technical University of Munich}
\mlsysaffiliation{ed}{Stanford University}

\mlsyscorrespondingauthor{Shu Liu}{lshu@berkeley.edu}

\mlsyskeywords{Machine Learning, MLSys}

\vskip 0.3in

\begin{abstract}

Batch data analytics is a growing application for Large Language Models (LLMs).
LLMs enable users to perform a wide range of natural language tasks, such as classification, entity extraction, and translation, over large datasets.
However, LLM inference is highly costly and slow: 
for example, an NVIDIA L4 GPU running Llama3-8B can only process 6 KB of text per second, taking about a day to handle 15 GB of data; processing a similar amount of data costs around \$10K on OpenAI's GPT-4o.
In this paper, we propose novel techniques that can significantly reduce the cost of LLM calls for relational data analytics workloads.
Our key contribution is developing efficient algorithms for reordering the rows and the fields within each row of an input table to maximize key-value (KV) cache reuse when performing LLM serving.  
As such, our approach can be easily applied to existing analytics systems and serving platforms. Our evaluation shows that our solution can yield up to 3.4$\times$ improvement in job completion time on a benchmark of diverse LLM-based queries using Llama 3 models. Our solution also achieves a 32\% cost savings under OpenAI and Anthropic pricing models. 

\end{abstract}

]



\printAffiliationsAndNotice{\mlsysEqualContribution} 

\section{Introduction}






One of the most popular applications of large language model (LLM) batch inference is data analytics. 
A growing number of analytics platforms now support LLM invocations for complex analytical tasks. 
For instance, leading database vendors, such as 
    AWS Redshift~\cite{aws-redshift-llm}, Databricks~\cite{databricks-ai-functions}, and Google BigQuery~\cite{google-bigquery-llm}, have integrated LLM functionality into their SQL APIs. 
    Similarly, DataFrame libraries and programming frameworks~\cite{langchain, lotus} offer LLM support for querying relational (table-based) data. 
With these new APIs, users can write queries like the following: 

\vspace{2pt}
\begin{mdframed}[linecolor=black, linewidth=.5pt]
\begin{minted}[fontsize=\small]{sql}
SELECT user_id, request, support_response, 
  LLM('Did {support_response} address {request}?', support_response, request) AS success
FROM customer_tickets 
WHERE support_response <> NULL
\end{minted}
\end{mdframed}
\vspace{-1pt}
where the LLM is invoked for each row in the customer ticket table to analyze whether the customer service requests are effectively addressed. 
Increasingly, analysts wish to leverage LLMs in such queries for tasks including classification, entity extraction, summarization, and translation~\cite{databricks-ai-functions}. Going forward, we will refer to queries that invoke LLMs over relational data as \textit{LLM queries}.

Unfortunately, applying LLMs to real-world datasets (which can contain millions of rows) incurs significant computational and monetary costs.       
Accordingly, there has been growing research on LLM inference optimization.
In particular, recent work~\cite{vllm, sglang, cascade-inference, hydragen, promptcache} leverages prompt caching, a technique that stores the attention states of frequently reused prompt segments in GPU memory, known as key-value (KV) cache ~\cite{attention-is-all-you-need}. Reusing cached state whenever a similar \textit{prefix} of prompts appears again can significantly reduce inference latency~\cite{sglang}. 
In addition, prompt reuse also brings economic benefits.
Recently, providers like OpenAI, Anthropic, and Google Gemini~\cite{openai-pricing, anthropicpromptcaching, gemini} have introduced prompt caching as a service, charging 2--10$\times$ less for cached prompts.  
Therefore, maximizing \textit{prefix hits} in the prompt KV cache is crucial for reducing both LLM request time and monetary costs.

However, simply invoking LLMs over relational data within analytical engines and connecting to a backend inference server with a prompt cache often results in low cache hit rates. This approach fails to exploit relational workloads to fully maximize cache reuse. 


In this work, we identify and present solutions to optimize relational data analytics workloads for offline LLM inference.
In particular, given an LLM query, we propose \textbf{request reordering} at the row and field granularity of the relational data. 
Our key insight is that, with oracular knowledge of all requests to be sent to the LLM, we can reorder both the requests and the fields inside each request to increase the number of cache hits. 
In real datasets, there can be many sharing opportunities across rows and fields. For example, joining feature tables, referencing popular items, or repeating similar context in RAG queries~\cite{retrieval-augmented-generation}.
These common patterns lead to repeating values in different fields, leaving rooms for significantly improving cache hit rates by optimizing request order and format. 

Finding the optimal ordering of requests is challenging due to the exponential number of choices to order the fields and rows of data in a query. For a table with $n$ rows and $m$ fields, there are $n! \times (m!)^n$ potential orderings. 
One way to reduce this search space is to apply the same field ordering across all rows. However, as we show in Sec~\ref{subsec:casestudy}, this can reduce the prefix hit count by up to a factor of $m$ compared to reordering fields on a more fine-grained, per-row basis.
To support per-row field reordering, we introduce \textbf{Optimal Prefix Hit Recursion (\optimal)}, an algorithm that divides the table into smaller subtables and reorders each subtable to maximize the prefix hits. While \optimal achieves high hit rates, its complexity is exponential, which makes it impractical for large datasets. To address this challenge, we propose \textbf{Greedy Group Recursion (\greedy)}, an approximate algorithm that leverages functional dependencies (such as primary and foreign key relationships from the data schema) and table statistics, which are readily available in many databases and analytics systems, to reduce the search space.
In particular, functional dependencies help identify correlated fields, reducing the number of fields that need to be reordered at each step, thus decreasing the solver runtime.
In addition, \greedy leverages the cardinality and length statistics to efficiently approximate the greedy objective. 

We implement our techniques in Apache Spark~\cite{zaharia2012resilient} and use vLLM~\cite{vllm} as the model serving backend. 
Due to the lack of standard workloads in this area, we build a benchmark suite of 16 LLM queries of different types, spanning selection, projection, multi-LLM invocations, and retrieval-augmented generation (RAG) queries~\cite{retrieval-augmented-generation}. We evaluate these queries on recommendation and question-answering datasets such as Amazon Product Reviews, Rotten Tomatoes Movies, BIRD, Stanford Question Answering Dataset, Public Domain MusicXML, RateBeer Reviews, and Fact Extraction and VERification datasets~\cite{amazon-product-review-dataset, rotten-tomatoes-movies-dataset, li2024can, squad-dataset, pdmx, fever}. Our techniques show 1.5 -- 3.4$\times$ speed-up in end-to-end query latency and reduce costs by up to 32\% on proprietary model APIs, while preserving query semantics. In summary, our contributions are as follows: 
\vspace{-0.5em}
\begin{itemize}
    \item We identify significant opportunities to speed up LLM-based batch data analytics through reordering rows and fields of input tables.  
    \item We introduce an optimal reordering algorithm (\optimal) that maximizes prefix sharing but with exponential complexity. 
    We propose an efficient greedy algorithm (\greedy) that approximates \optimal by leveraging functional dependencies and table statistics. We show that a fixed field ordering can yield as much as $m$ (number of fields) times worse cache hits than our solution.
    \item We present an LLM query benchmark consisting of 16 queries and 7 real-world datasets to represent a range of retrieval and processing tasks. Our evaluation with Llama3-8B and 70B shows up to a 3.4$\times$ speedup in end-to-end query latency compared to naive orderings. With OpenAI and Anthropic prefix cache pricing models, our techniques reduce costs by up to 32\%.
\end{itemize}

\vspace{-1em}

\section{Background and Motivation}
\label{sec:motivation}
This section provides a brief overview of the inference process and the key components of the LLM architecture.

\textbf{LLM inference.} 
LLMs are made up of autoregressive Transformer models~\cite{attention-is-all-you-need}, which generate text token by token until a termination token or a length limit is reached. LLM inference consists of two stages: (i) the prefill stage, where the model processes the input prompts, and (ii) the decoding stage, where it generates output sequentially, as each token depends on all previously generated tokens through a chain of conditional probabilities.
LLM inference engines (e.g., vLLM \cite{vllm}, TGI \cite{tgi}, TensorRT-LLM \cite{trt-llm}) typically batch requests continuously \cite{orca-continous-batching} to improve throughput. 
The intermediate computed state for all tokens involved is stored in memory.
This token state is cached as key and value vectors in the \textit{key-value (KV) cache}, consuming up to 800KB per token for a 13B Model \cite{vllm}. 
A typical request (involving 2,000 tokens) can require up to 1.6 GB of memory. 
Despite batching (up to 32 requests), inference remains compute-intensive, with current speeds limited to ~2,000 tokens/s per GPU, making LLM performance a bottleneck for many analytical tasks.

\textbf{Prompt KV cache.}
Efficient KV cache management is critical for high LLM serving throughput.
Recent work improves cache utilization by reusing tokens across requests with shared prefixes~\cite{sglang}.
For example, if two requests share a \textit{prefix} in prompts, the first will already have performed some computation on the input tokens and cached results in the KV cache during the prefill phase. 
The subsequent request can then reuse these cached values, avoiding redundant computation of the shared tokens.

\textbf{Improving KV cache hit for analytics workloads}.
Real-world relational databases often exhibit diverse repetitive data patterns. 
Columnar storage systems like C-Store and Parquet~\cite{stonebraker2018c} exploit repeated values across fields for compression, while techniques like run-length encoding (RLE), multi-relational data mining, and correlation analysis~\cite{lemire2011reordering, multirelation,correlation} leverage diverse data relationships to optimize query execution. 
Relational queries also create data groupings based on access patterns. 
Techniques such as database cracking and multi-dimensional clustering (MDC)~\cite{craking,mdc}, including Delta Lake Z-order~\cite{deltalake}, reorganize data based on query patterns to optimize performance.

These structural repetitive patterns present an opportunity for \textit{prefix KV cache} sharing in an LLM query.
In our setting, an LLM is invoked once per row in a relational table, resulting in a batch of model requests from a single LLM query. Since the full table structure and content are known in advance, we can reorder these requests to maximize shared prefixes and reduce redundant computation during inference. Our goal is to maximize the \textit{prefix hit count} -- the sum of the length of token prefixes reused from the KV cache. 




\textit{Our Approach: Request Reordering.} We leverage table information to enhance the KV cache hit rate. Specifically, we introduce algorithms that reorder requests of an LLM query and fields within each request to maximize prefix sharing. Our algorithm leverages functional dependencies and table statistics to reduce runtime while finding near-optimal orderings that maximize prefix hit count.


%
\vspace{-0.8em}
\section{Problem Setup}
This section introduces the problem setup of maximizing prefix hits in the prompt cache (Sec~\ref{subsec:setup}) and highlights cases where naive fixed field ordering can result in significantly lower hit rates (Sec~\ref{subsec:casestudy}).

\begin{figure*}[tbp]
     \centering
     \begin{subfigure}[b]{0.48\textwidth}
        \centering
        \includegraphics[width=\textwidth]{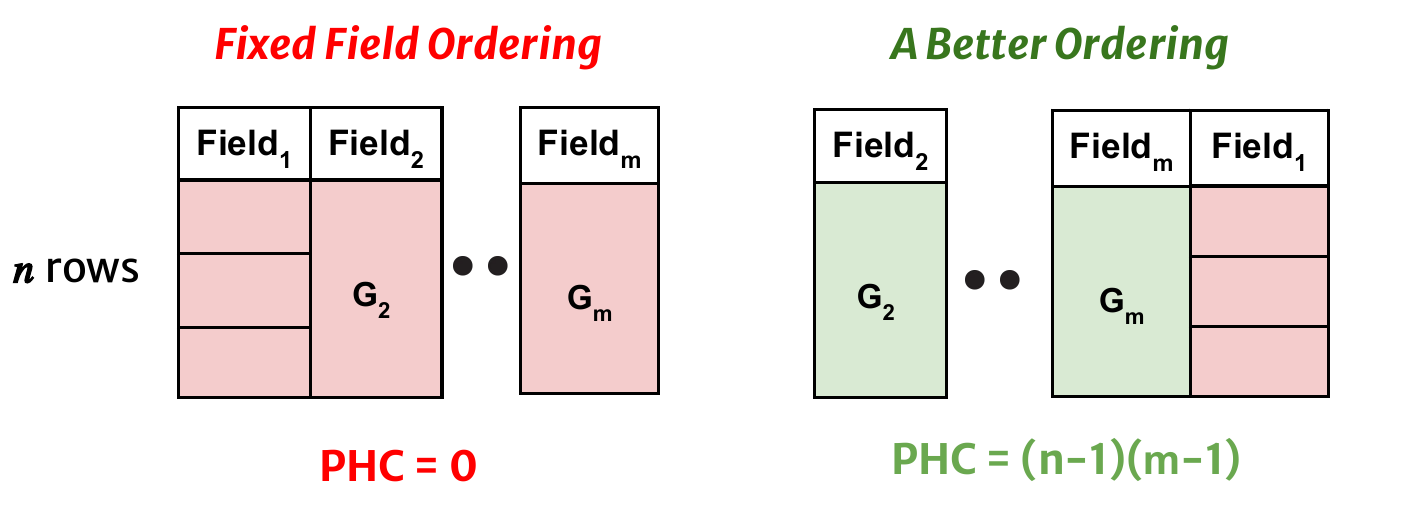}
        \caption{Distinct Values in the First Field}
        \label{fig:cases-1}
    \end{subfigure}
    \hfill
    \begin{subfigure}[b]{0.48\textwidth}
        \centering
        \includegraphics[width=\textwidth]{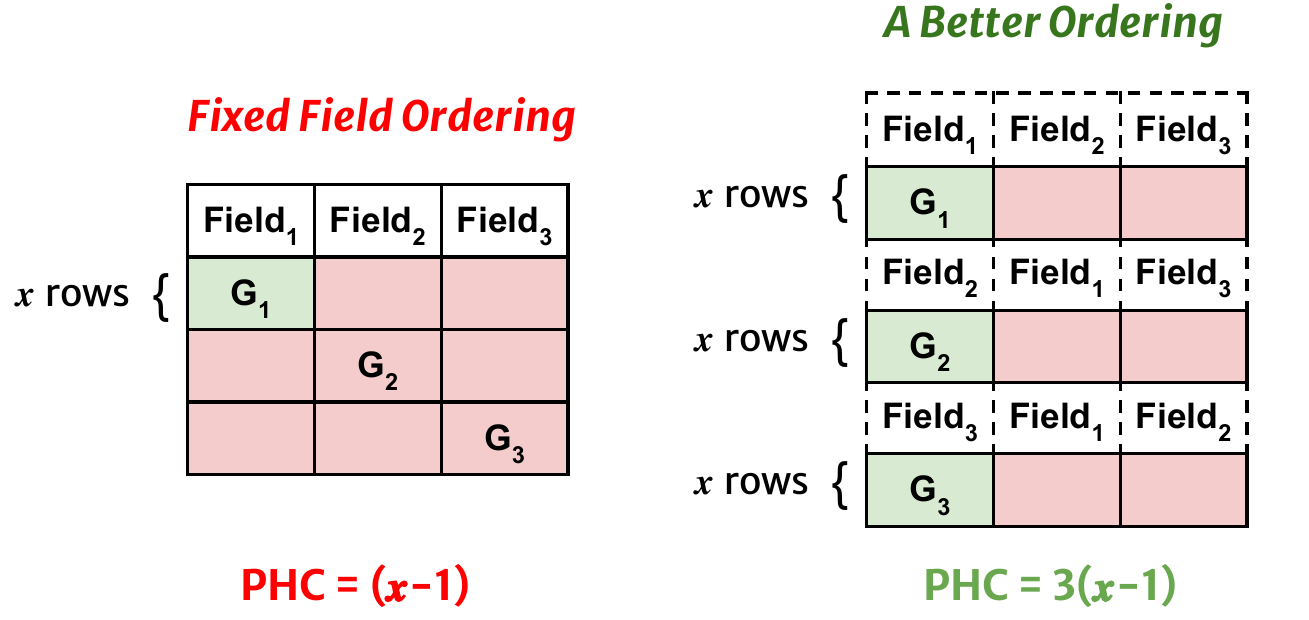}
        \caption{Group of Identical Values in each Field, $m = 3$}
        \label{fig:cases-2}
    \end{subfigure}

    \vspace{-0.5em}
    \caption{\textbf{Case Study of Fixed Field Ordering:} Comparing the PHC of a fixed field ordering to a better ordering in two scenarios. Green boxes denote cache hits; red boxes indicate cache misses. A box labeled $G_{i}$ signifies consecutive rows share the same values in Field $i$; otherwise, assume values are distinct. Fig~\ref{fig:cases-1} shows fixed field ordering can be $(n-1)(m-1)$ worse in terms of PHC compared to an optimized ordering. Fig~\ref{fig:cases-2} shows fixed field ordering can be $m$ times worse in PHC compared to an optimized ordering, where $m = 3$.}
    \label{fig:runtimes}
    \vspace{-1.5em}
\end{figure*}
\vspace{-0.5em}
\subsection{Setup and Objective}
\label{subsec:setup}
In this work, we consider a generic LLM operator that takes the text of the prompt as well as a \emph{set} of expressions listing one or more fields $\{T.a, T.b, T.c\}$ or $\{T.*\}$ of the table $T$. This simple design can be easily implemented in most analytics systems and enables us to dynamically reorder fields within these expressions to optimize for cache efficiency. Consider the following example query:

\vspace{-0.13em}
\begin{mdframed}[linecolor=black, linewidth=.5pt]
\begin{minted}[fontsize=\small]{sql}
SELECT LLM("Summarize: ", pr.*) 
FROM (
    SELECT review, rating, description
    FROM reviews r JOIN product p ON r.asin = p.asin
) AS pr
\end{minted}
\end{mdframed} 
\vspace{-0.2em}

This query sends a list of rows, each with fields \textit{review}, \textit{rating}, and \textit{description} from table \textit{pr} to the LLM for a summarization task. 


\textbf{Objective} The goal of request scheduling is to \textbf{maximize} the \textit{prefix hit count} by optimizing the order of fields and rows of an input table with $n$ rows and $m$ fields. 
Each row may have a different field order. 
We represent a request schedule as a list of tuples $L$, where each tuple in $L$ represents a row in the table, and the tuple elements contain the field values. 
We adjust the row order by rearranging the tuples in $L$, and adjust the field order for that row by rearranging the elements within each tuple. 
We pass each tuple alongside the user question to form an input request to the LLM. 


We define the \textit{prefix hit count} (PHC) of $L$ as the number of consecutive field cell values shared with the previous row starting from the first cell, summing over all $n$ rows. 
Each cell value must exactly match the corresponding cell of the previous row (cannot be a substring), and cell values past the first must match consecutively (must be a prefix). 
Formally, a cell in the list of tuples is denoted as $L[r][f]$, indicating the value in tuple $r$ at position $f$. Then, the PHC for a list of tuples $L$ with $n$ rows and $m$ fields is given by: 
\vspace{-1.5em}

\begin{equation}
\text{PHC}(L) = \sum_{r=1}^{n} \textit{hit}(L, r)
\label{eq:phc}
\end{equation}
\vspace{-1.5em}

Here, the function $\textit{hit}(L, r)$ represents the prefix hit count for a single row $r$ in $L$. For simplicity, we assume that the input list is sorted. For each row $r$, the function checks if the value in each field $f$ matches any previously seen value in the same field of the previous row $r-1$. If all previous fields match, the hit count is the sum of the squares of the lengths of the values in those fields until a mismatch occurs. The squared lengths reflect the quadratic complexity of token processing in LLM inference, where each token computation depends on every preceding token and increases computational cost quadratically with input length. 


\vspace{-2em}

\begin{equation}
\textit{hit}(L, r) = \max_{0 \leq c < m}
\begin{cases} 
\sum_{f=1}^{c} \text{len}(L[r][f])^2 & \text{if } \forall f \leq c, \\ & L[r][f]= \\ & L[r-1][f] \\

0 & \text{otherwise}
\end{cases}
\label{eq:hit}
\end{equation}
\vspace{-1.5em}

To simplify the design, we make two assumptions. 
First, we make a common assumption that at least one tuple (row) can fit into the KV cache to allow reuse.
Second, we assume that a cell value only counts as a hit if it exactly matches a previously seen value -- substring matches are not allowed. 
This is a reasonable assumption in relational databases, where exact value repetition is common and extensively leveraged by storage optimization techniques like run-length encoding~\cite{lemire2011reordering}. Column-oriented storage systems such as C-Store and Parquet~\cite{stonebraker2018c} also benefit from many exact repetitions in columnar data.
These assumptions simplify design and, as shown in Sec.~\ref{sec:evaluation}, demonstrate good real-world performance.


\subsection{Case Study: Fixed Field Ordering}
\label{subsec:casestudy}
Relational data typically uses a fixed field order across rows, which can lead to lower hit rates in real-world databases with diverse data patterns (Sec~\ref{sec:motivation}).
In fact, we show that using a fixed order can reduce the hit rate by up to $m$ times compared to a per-row field reordering.
To illustrate this, we begin with a simple example and extend it to show the potential impact of a naive fixed field ordering on prefix hit counts (PHC). 
First, consider a table $T$ with $n$ rows and $m$ fields arranged in an arbitrary (default) order. 
For simplicity, we assume each value is of length one.
In many cases, certain fields of an input table may contain highly unique values, like timestamps or IDs. 
In the worst case, suppose the first field of the table contains only unique values (Fig~\ref{fig:cases-1}), and the remaining $m-1$ fields contain the same value across all rows. 
This ordering yields $0$ PHC. 
A more optimized ordering (Fig~\ref{fig:cases-1}) will place the other $m-1$ fields first, yielding a PHC of $(n-1) \times (m-1)$. Each of the $n-1$ rows has a hit after the initial cold miss, and the length of each hit is $m-1$. 

Now consider a scenario where the table contains groups of consecutive rows with identical values (not necessarily in the same field). 
Suppose each field $i$ has one such group with $x$ consecutive rows of the same value, with other $n-x$ rows having distinct values, where $n$ is the number of rows. 
We denote the group appearing in the $\text{Field}_i$ as $G_i$, so we have $G1, ..., G_{m}$ groups, where $m$ is the number of fields. 
Now, consider a scenario where groups in consecutive fields are non-overlapping across rows, as shown in Fig~\ref{fig:cases-2}. 
With fixed field reordering, the PHC of this structure is limited to $x-1$ no matter which field is prioritized. 
By contrast, a better ordering would rearrange the field order for different rows to prioritize groups with shared values. 
Fig~\ref{fig:cases-2} references a table with $3x$ rows and 3 fields. A naive fixed field ordering for all rows will result in misses on two groups, each with $x$ rows in $\text{Field}_{2}$ and $\text{Field}_{3}$. However, a better ordering will pick different $\text{Field}_{j}$ to prioritize for different rows, resulting in a 3 times higher hit rate of $3(x-1)$.

In the above scenario, PHC improvements from optimized field ordering can reach $m$ times that of a fixed field ordering. For example, there can be multiple (instead of just one) such groups in each field. 
If each field contains roughly the same number of such groups, dynamic reordering for different rows can achieve as much as an $m$-fold improvement in PHC over fixed field ordering. 
Under the OpenAI pricing model, which charges half price for cached prompts, optimizing field order for a table with nine fields could yield 42\% in cost savings compared to fixed field ordering, assuming fixed ordering has a 10\% hit rate (e.g., $\frac{(x-1)}{n} = 10$). 
This example highlights the benefits of a more complex field reordering mechanism for different rows on PHC.



\vspace{-0.5em}
\section{Recursive Request Reordering}
\label{sec:column-reordering}


We now introduce our algorithms that re-arrange fields to maximize prefix sharing in the KV cache. 
We present an optimal recursive reordering algorithm that maximizes PHC (Sec~\ref{sec:optimal}) and 
introduce a greedy algorithm that efficiently approximates the optimal algorithm  (Sec~\ref{sec:greedy}).

\begin{algorithm}[t!]
\caption{Greedy Group Recursion (GGR)}
\begin{algorithmic}[1]
\small
\STATE \textbf{Input:} Table $T$, Functional Dependency $FD$
\STATE \textbf{Output:} Prefix Hit Count $S$, Reordered List of Tuples $L$

\item[]
\FUNCTION{$\textsc{HitCount} (v, c, T, FD)$}
    \STATE $R_v \gets \{i \mid T[i,c] = v\}$
    \STATE $\text{inferred\_cols} \gets \{c' \mid (c, c') \in FD\}$
    \STATE $\text{tot\_len} = \text{len}(v)^2 + \sum_{\substack{c' \in \text{inferred\_cols}}} \frac{\sum_{r \in R_v} \text{len}(T[r, c'])}{|R_v|}$
    \STATE \textbf{return } $\text{tot\_len} \times (|R_v| - 1)$, $[c] + \text{inferred\_cols}$
\ENDFUNCTION

\item[]
\FUNCTION{\textsc{GGR}($T$, $FD$)}
    \IF{$|T|_{rows} = 1$}
        \STATE return 0, $[T[1]]$
    \ENDIF 
    \IF{$|T|_{cols} = 1$}
        \STATE $S \gets \sum_{v \in \text{distinct}(T[,1])} \textsc{HitCount}(v, 1, T)$ 
        \STATE {\bfseries Return} $S, sort([T[i] \mid i \in 1 \dots |T|_{rows}])$
    \ENDIF

    \STATE $max\_HC, b\_v, b\_c, b\_cols \gets -1, \text{None}, \text{None}, []$

    \FOR{$c \in \text{columns}(T)$, $v \in \text{distinct}(T[,c])$}
        \STATE $HC, cols \gets \textsc{HitCount}(v, c, T, FD)$
        \IF{$HC > max\_HC$}
            \STATE $max\_HC, b\_v, b\_c, b\_cols = HC, v, c, cols$
        \ENDIF
    \ENDFOR
    
    \STATE $R\_v \gets \{i \mid T[i, b\_c] = b\_v\}$
    \STATE $A\_HC, L\_A \gets \textsc{GGR}(T[\text{rows} \setminus R\_v, \text{cols}], FD)$
    \STATE $B\_HC, L\_B \gets \textsc{GGR}(T[R\_v, \text{cols} \setminus b\_cols], FD)$
    \STATE $C\_HC, \_ \gets \textsc{HitCount}(b\_v, b\_c, T, FD)$
    \STATE $S \gets A\_HC + B\_HC + C\_HC$
    \STATE $L \gets [[b\_v] + L_A[i] \mid i \in 1 \dots |R\_v|] + L\_B$
    \STATE \textbf{return } $S, L$
\ENDFUNCTION

\item[]
\STATE \textbf{return } \textsc{GGR}($T$, $FD$)
\end{algorithmic}
\label{alg:greedy}
\end{algorithm}
\vspace{-0.5em}
\subsection{Optimal Prefix Hit Recursion (OPHR) } \label{sec:optimal}





Our Optimal Prefix Hit Maximization (\optimal) algorithm is a recursive algorithm that finds the \textit{optimal} PHC for a given table $T$ by considering all possible ways to split the table into a group of cells with the same value and two sub-tables. 
The algorithm takes as input a table $T$ and computes the optimal PHC $S$ along with a reordered list of tuples $L$. 
If $T$ only has one row or field, \optimal computes PHC and trivially returns the sorted $T$. 

In the recursive case, for each field $c$ in $T$, the algorithm identifies all distinct values $v$ in the field and the rows $R_v$ for which the field value is $v$. For each distinct value $v$, the table is split into two sub-tables: one of $T$ excluding rows $R_v$ and one of $R_v$ excluding field $c$. PHC for the currently selected value $v$ is calculated as the sum of the PHC of the sub-tables and the PHC contribution of $v$. \optimal evaluates all possible groups of distinct values in each field and selects the value that yields the maximum PHC. 


Notably, the \optimal algorithm has exponential complexity with respect to the number of rows and fields due to its recursive nature and the combinatorial explosion of possible distinct value groupings (we present a more efficient algorithm in Sec~\ref{sec:greedy}).

\begin{figure}
    \centering
    \includegraphics[width=0.49\textwidth]{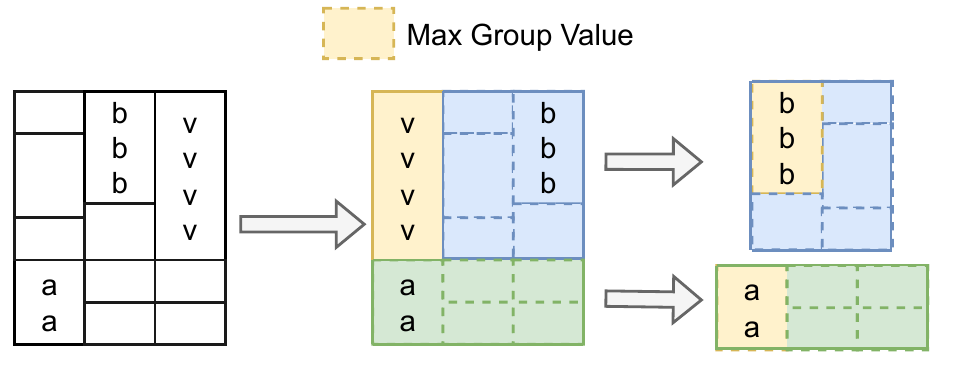}
    \vspace{-1.5em}
    \caption{GGR picks the group with the maximum hit count at each step and calculates PHC as the sum of PHC of the elected group values (yellow box), the sub-table $T$ excluding rows $R_v$ (green box), and the sub-table of rows $R_v$ excluding the field where the value is located in (blue box).}
    \label{fig:optimal}
    \vspace{-2em}
\end{figure}

\textbf{Optimality Proof}
In the base case, the \optimal algorithm trivially computes the best PHC: for the single row case, the PHC is 0; for the single field case, the PHC is the sum of the squared lengths of distinct values multiplied by their occurrences minus one, which accounts for the initial miss when a value is seen the first time. 
Next, we prove optimality by induction.
For the inductive case, assume that the \optimal algorithm is optimal for any table with $k \leq n$ rows and $l \leq m$ fields. 
For a table $T$ with $n+1$ rows and $m+1$ fields, the algorithm iterates through each field $c$. For each distinct value $v$ in field $c$, we split $T$ into two sub-tables: $T_A$ (rows not containing $v$), and $T_B$ (rows containing $v$ but excluding field $c$). Based on the inductive hypothesis, \optimal optimally computes PHC for both sub-tables because it is optimal for tables with fewer rows and fields. The PHC for $T$ is the sum of PHC for $T_A$ and $T_B$, plus the contribution of $v$. When the distinct value $v$ is used to partition the table, its full contribution to the PHC is captured. If the table were not split based on distinct values, this contribution could be fragmented or lost due to non-contiguous groupings, leading to suboptimal PHC.
Thus, the \optimal algorithm ensures optimal reordering by selecting the best from all possible configurations.
\vspace{-0.5em}
\subsection{Greedy Group Recursion (\greedy) Algorithm}
\label{sec:greedy}
Due to the computational complexity of the \optimal, we propose a Greedy Group Recursion (\greedy) algorithm (Algorithm~\ref{alg:greedy}) that approximates \optimal.  
The \greedy algorithm takes an input table $T$ and returns the PHC $S$ along with a reordered list of tuples $L$. It has the same base case as the \optimal algorithm if $T$ only has one row or one field. At a high level, the \greedy algorithm recursively selects the value $b_v$ with the maximum prefix hit count (lines 3-8) at each recursion step (lines 17-23) rather than iterating through all possible distinct values in the entire table. 
It then prioritizes the field $b_c$ where this $b_v$ is in, splits the table into groups of cells of the same values and recurses on the two sub-tables (lines 24-26) and calculates the total PHC as the sum of PHC of the subtables and contributions of $b_v$ (line 28) similar to the \optimal algorithm. 

Since \greedy does not iterate through all possible distinct values but instead selects the one that gives the highest hit count at each step, the number of recursive calls is significantly reduced (i.e. the maximum depth of recursion is $O(\min(n, m))$, where the algorithm reduces dimensions of the table at each recursive step). However, at each recursive step, the cost of scanning to determine distinct values can result in quadratic complexity in terms of table size.

\vspace{-0.5em}
\subsubsection{Functional Dependencies} We leverage functional dependencies to reduce the number of fields the \greedy algorithm needs to consider at each recursion step. This insight helps improve both the approximation and efficiency of the algorithm, bringing it closer to the optimal solution without the need for extensive backtracking as in the \optimal algorithm. 
A functional dependency (FD) is a constraint between two sets of attributes in a relation from the data. For example, let $R$ be a relation schema and let $X$ and $Y$ be nonempty sets of attributes in $R$. We define an instance $r$ of $R$ that satisfies the FD $X \leftrightarrow Y$ if for every pair of tuples $t_1$ and $t_2$ in $r$: if $t_1.X = t_2.X$ then $t_1.Y = t_2.Y$ and vice versa. In our \greedy algorithm, FDs help narrow down the fields that must be considered at each recursion step. Specifically, when a value $v$ in field $f$ is selected for a given row, all fields functionally dependent on $f$ are ordered directly besides $f$ in the final ordering for that row (lines 5-6). As an example, if $R(A,B,C)$ is a table with attributes (fields) $A,B,C$ where we have an FD $A \leftrightarrow C$, field $C$ is not in consideration in our recursive steps when $A$ has already been included in the prefix.
\vspace{-0.5em}

\subsubsection{Table Statistics} 
To further reduce the algorithm runtime, we introduce an early stopping mechanism that halts recursion by specific recursion depth (row-wise sub-table recursion, column-wise sub-table recursion) or when a threshold $\mathtt{HITCOUNT}$ score calculated using table statistics is not exceeded. These statistics are generally widely available, such as the number of unique entries (i.e., cardinality) and the distribution of length of values for each field.
With this information, our \greedy algorithm estimates a $\mathtt{HITCOUNT}$ score for each field $c$ with $\mathtt{HITCOUNT}(C) = \mathtt{avg}(\mathtt{len}(c))^2$. This score denotes the expected contribution of a field to the PHC, accounting for the average length of the values and their frequency. Using these statistics, the algorithm can prioritize fields more likely to contribute to the PHC. 
Additionally, we can further improve the quality of the solution by establishing a fixed field ordering for the subtables using table statistics once the recursion stops. 
Early termination and falling back to table statistics allows \greedy to avoid scanning the table and performing recursion on real-world workloads. 

\vspace{-0.5em}
\subsubsection{Achieving Optimal PHC} While our \greedy approximates the \optimal algorithm, it can achieve optimal PHC in certain cases. 
When the table has only one row or one field, \greedy matches \optimal by construction. 
When functional dependencies are accurate and cover all the fields of a table, \greedy can also identify the optimal solution. For instance, if one field $A$ functionally determines all other fields, then \greedy prioritize groups of values in $A$ due to the accumulated \texttt{HITCOUNT} score (line 3 in Algorithm~\ref{alg:greedy}), capturing key correlations early and producing the optimal reordering.
However, when fields tie in \texttt{HITCOUNT}, \greedy may be suboptimal, as it lacks the exhaustive search used by \optimal to resolve such ties. 
We show more empirical results in real-world datasets comparing PHC between \greedy and \optimal in Appendix~\ref{appendix:hit-rate}.

\vspace{-0.5em}

\section{Implementation}
We implement our algorithms in approximately 1.3K lines of Python code and evaluate them with PySpark~\cite{spark-sql}, which is backed by Apache Spark~\cite{zaharia2012resilient} -- a widely adopted large-scale data processing engine in industry.
The \textit{LLM operator} implements the actual
LLM inference by calling a configurable LLM endpoint. We implement this function as a UDF in PySpark. It takes in a system prompt, a query prompt, and a single row of data as input (Appendix~\ref{appendix:prompts}). 
The row and field orders are input based on the ordering returned by the reordering function. 
The operator is also responsible for \textit{prompt construction}. Specifically, it converts the user-provided question and the table row values into a prompt that an LLM can parse. We use JSON formatting to encode row values to indicate the relationship between field names and values to the LLM. 
\vspace{-0.5em}

\section{Evaluation}
\label{sec:evaluation}
In this section, we evaluate the effectiveness of our optimizations within a constructed benchmark suite of queries. We aim to answer the following questions: 
\vspace{-1em}

\begin{enumerate}\setlength{\itemsep}{-1.5pt}
    \item How does our request reordering optimization impact query latency and costs across different LLM query types and datasets? 
    \item How does the request reordering algorithm influence LLM accuracy for different models?
    \item What is our algorithm solver time, and how does that compare to end-to-end query latency? 
\end{enumerate}

\vspace{-1em}

\subsection{Evaluation Benchmark}
\label{sec:queries}
Given the lack of standard benchmarks for LLM queries, we construct a benchmark suite to represent real-world data retrieval and processing tasks (Sec~\ref{subsec:dataset}). 
We define a range of query types (Sec~\ref{subsec:llmqueries}) over datasets from various sources to assess the impact of LLMs in relational analytics.

\subsubsection{Datasets}
\label{subsec:dataset}
\begin{table}[ht]
\centering
\small
\begin{tabular*}{\columnwidth}{@{\extracolsep{\fill}}cccccl}
\toprule
Dataset & \( n_{\text{rows}} \) & \( n_{\text{fields}} \) & \( \text{input}_{\text{avg}} \)  & \( \text{output}_{\text{avg}} \) & \text{Query Type} \\ 
\midrule
Movies & 15000 & 8 & 276 & $\{2, 29, 16, 2\}$ & T1-T4 \\ 
Products & 14890 & 8 & 377 & $\{3, 107, 62, 2\}$ & T1-T4\\
BIRD & 14920 & 4 & 765 & $\{2, 43\}$ & T1, T2 \\ 
PDMX & 10000 & 57 & 738 & $\{2, 72\}$ & T1, T2\\
Beer & 28479 & 8 & 156 & $\{2, 38\}$ & T1, T2\\ 
SQuAD & 22665 & 5 & 1047 & {11} & T5\\
FEVER & 19929 & 5 & 1302 & {3} & T5\\ 
\bottomrule
\end{tabular*}
\caption{Datasets: $n_{\text{rows}}$ and $n_{\text{fields}}$ denote the number of rows and fields, respectively. $\text{input}_{\text{avg}}$ and $\text{output}_{\text{avg}}$ represent average input and output token lengths. Query Type is detailed in Sec~\ref{subsec:llmqueries}. Since $\text{input}_{\text{avg}}$ remains consistent across query types, we report a single overall average, while $\text{output}_{\text{avg}}$ varies, with each bracketed value corresponding to a specific query type.} 
\label{tab:dataset}
\vspace{-1em}
\end{table}



We build our benchmark suite on 7 commonly used recommendation and natural language processing datasets, shown in Table~\ref{tab:dataset}. These datasets vary in the number of rows, fields, average input/output token lengths, and appropriate query types (Sec~\ref{subsec:llmqueries}). 
The datasets include Rotten Tomatoes Movie Reviews (Movies)~\cite{rotten-tomatoes-movies-dataset}, Amazon Product Reviews (Products)~\cite{amazon-product-review-dataset}, BIRD~\cite{li2024can}\footnote{We use Posts and Comments table joined by PostID from the BIRD dataset.}, Public Domain MusicXML (PDMX)~\cite{pdmx}, RateBeer Reviews (Beer)~\cite{ratebeer}, Stanford Question Answering Dataset (SQuAD)~\cite{squad-dataset}, and Fact Extraction and Verification (FEVER)~\cite{fever}. Details on the fields are in the Appendix~\ref{appendix:fields}.

\vspace{-0.5em}

\subsubsection{LLM Queries}\label{subsec:llmqueries}

Our evaluation consists of 16 queries across 5 query types corresponding to different real-world use cases, as shown in Table~\ref{tab:dataset}. 
We discuss each query type below and provide details on queries for each dataset in Appendix~\ref{appendix:queries} and ~\ref{appendix:fields}. 


\textbf{\textit{(T1) LLM filter.}} Filter queries mimic SQL \texttt{WHERE} clauses and use LLMs to categorize data. This query type illustrates typical use cases in sentiment analysis, categorization, and content filtering.
Given their binary or categorical focus, these queries often yield short outputs (e.g., "Yes" or "No"). We construct five filter queries spanning all datasets except for SQuAD and FEVER. \newline 
\textbf{\textit{(T2) LLM projection.}} Projection queries use LLMs to summarize or interpret specific table field(s), similar to a SQL \texttt{SELECT} statement. 
These tasks typically produce longer outputs due to the descriptive nature of the results. We construct five projection queries spanning all datasets except SQuAD and FEVER. \newline 
\textbf{\textit{(T3) Multi-LLM invocation.}} Multi-LLM queries involve sequential LLM calls (e.g., a filter followed by a projection), supporting tasks like multi-step data processing and combining insights. 
Output lengths vary by task but generally mix short and long responses.
We construct two example multi-LLM invocation queries on Movies and Products datasets. \newline 
\textbf{\textit{(T4) LLM aggregation.}} 
Aggregation queries incorporate LLM outputs into aggregate functions, like averaging sentiment scores given by LLMs for individual reviews. These tasks usually generate concise numeric outputs for analysis (e.g., ratings of 1 to 5), resulting in shorter output lengths similar to filter queries. We construct two example aggregation queries on Movies and Products datasets. \newline 
\textbf{\textit{(T5) Retrieval-augmented generation (RAG)}.} RAG queries involve fetching external knowledge as context, such as retrieving relevant document segments before generating answers. We evaluate FEVER and SQuAD datasets, fetching 4 contexts for FEVER and 5 contexts for SQuAD for question answering.



\begin{figure*}[tbp]
     \centering
     \begin{subfigure}[b]{0.48\textwidth}
        \centering
        \includegraphics[width=\textwidth]{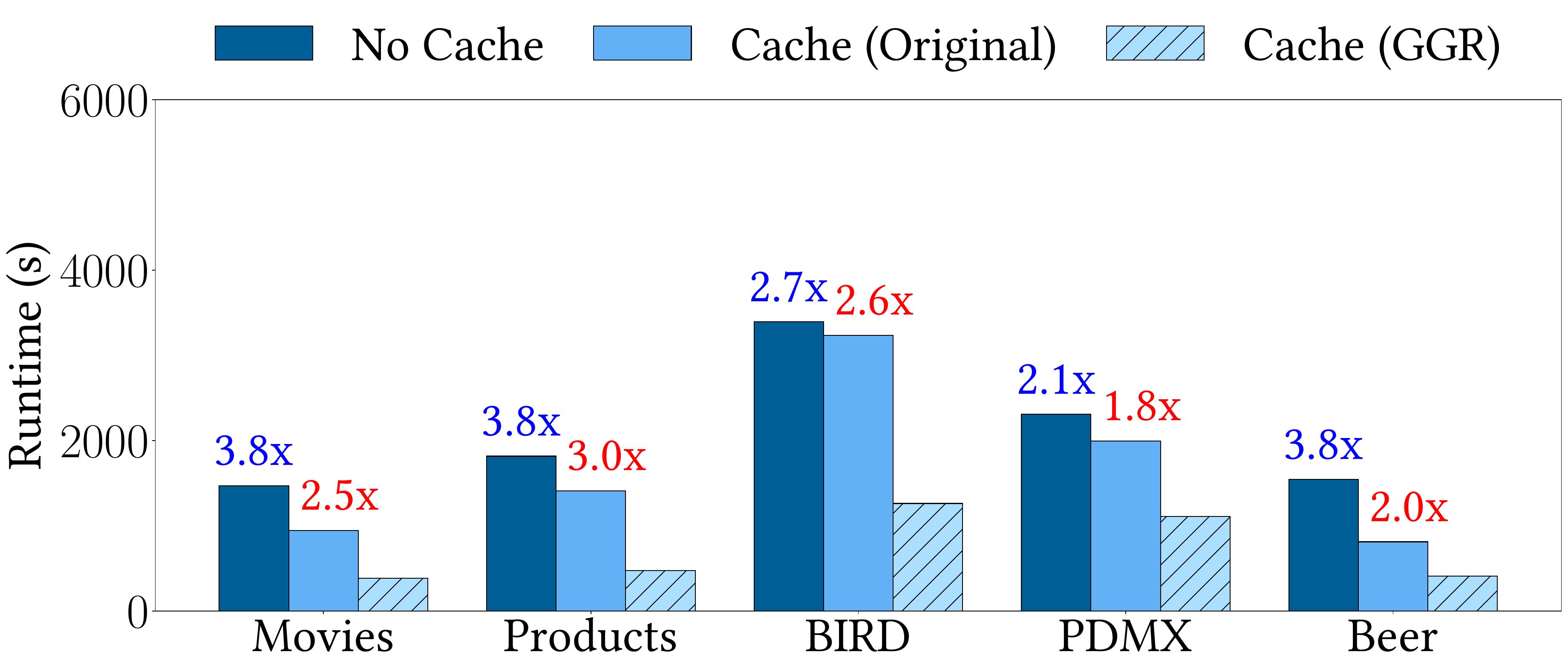}
        \caption{Filter Queries}
        \label{fig:filter-q}
    \end{subfigure}
    \hfill 
    \begin{subfigure}[b]{0.48\textwidth}
        \centering
        \includegraphics[width=\textwidth]{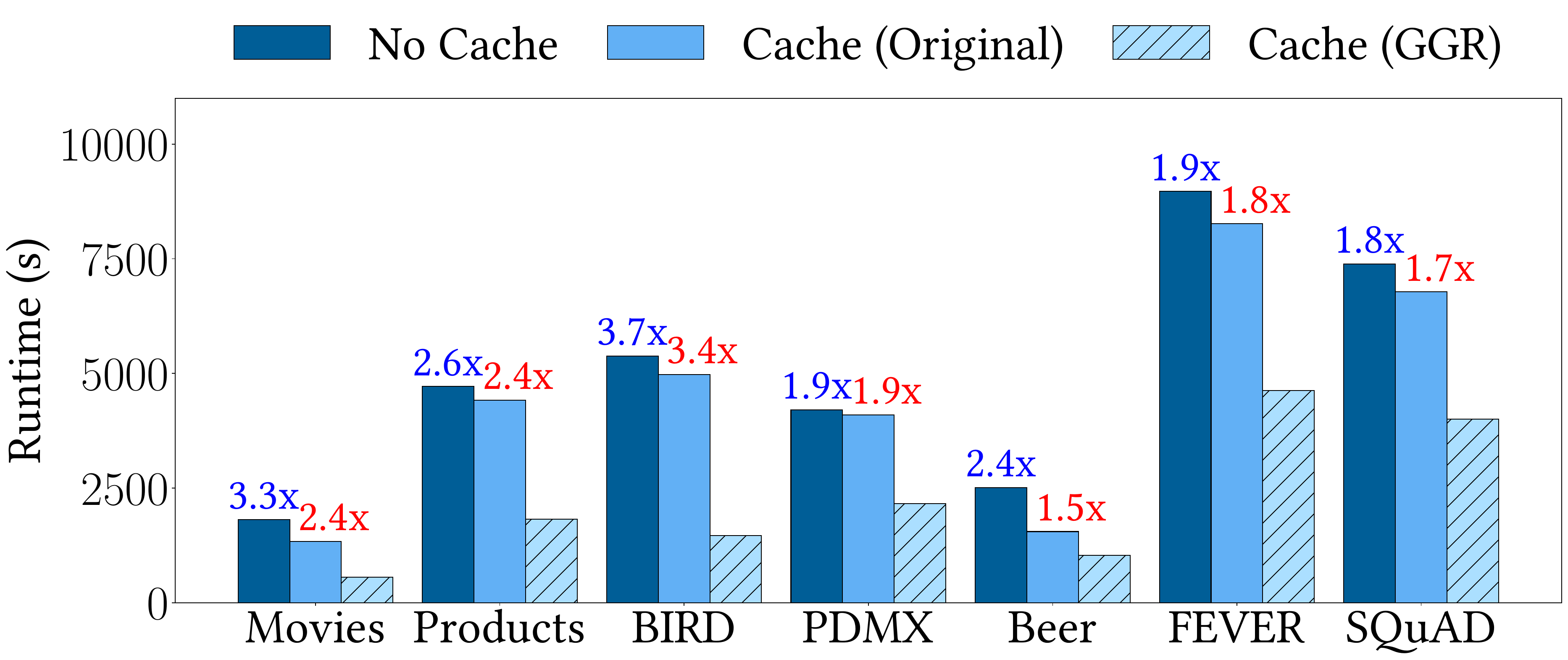}
        \caption{Projection and RAG Queries}
        \label{fig:selection-rag}
    \end{subfigure}
    \vspace{-1em}
    \caption{End-to-end Result (Filter, Selection, RAG): Our optimizations Cache (\greedy) achieve 1.5 -- 3.4$\times$ speed-up in end-to-end runtime over caching without reordering (Cache (Original)), and 1.8 -- 3.8$\times$ over No Cache baseline. }
    \label{fig:q1q5}
    \vspace{-1.3em}
\end{figure*}









\vspace{-0.5em}
\subsubsection{Evaluation Setup}

\textbf{Metrics} We evaluate \textit{end-to-end query latency} for each LLM query. We also measure the \textit{monetary cost} of using OpenAI and Anthropic endpoints. Additionally, we hand-label a subset of the LLM filter queries to evaluate the reordering implications for query \textit{accuracy}. \newline 
\textbf{Models} We run setups shown in Table~\ref{tab:dataset} using Meta Llama-3-8B-Instruct ~\cite{llama3}. For RAG queries, we use Alibaba-NLP/gte-base-en-v1.5~\cite{li2023towards} to embed the context and use Facebook Similarity Search Library (FAISS)~\cite{johnson2019billion} for context retrieval. We also run Llama-3-70B-Instruct~\cite{llama3} for LLM Filter queries. For cost results, we evaluate with OpenAI GPT-4o-mini and Anthropic Claude 3.5 Sonnet.  \newline 
\textbf{Hardware} We evaluate Llama-3-8B-Instruct on a single NVIDIA L4 GPU (GCP g2-standard-4) with 24GB of GPU Memory. We also run a larger model Llama-3-70B-Instruct on 8xL4 GPUs (GCP g2-standard-48). For OpenAI and Anthropic cost experiments, we utilize their API endpoints. \newline 
\textbf{Baselines} Our algorithm (\textit{Cache (GGR)}) is compared against two baselines: one without prompt caching (\textit{No Cache}) and another with caching enabled but without reordering (\textit{Cache (Original)}). We do not evaluate the optimal prefix hit recursion algorithm (Sec~\ref{sec:optimal}) as it is infeasible over large tables (e.g., solving a 10-row table takes several minutes). 
The algorithm runtime far exceeds the LLM inference time for larger tables for the optimal algorithm. 

\vspace{-0.5em}

\subsection{End-to-End Benchmark Results}
\label{sec:end-to-end}

\begin{figure}[t!]
    \centering
    \includegraphics[width=0.95\columnwidth]{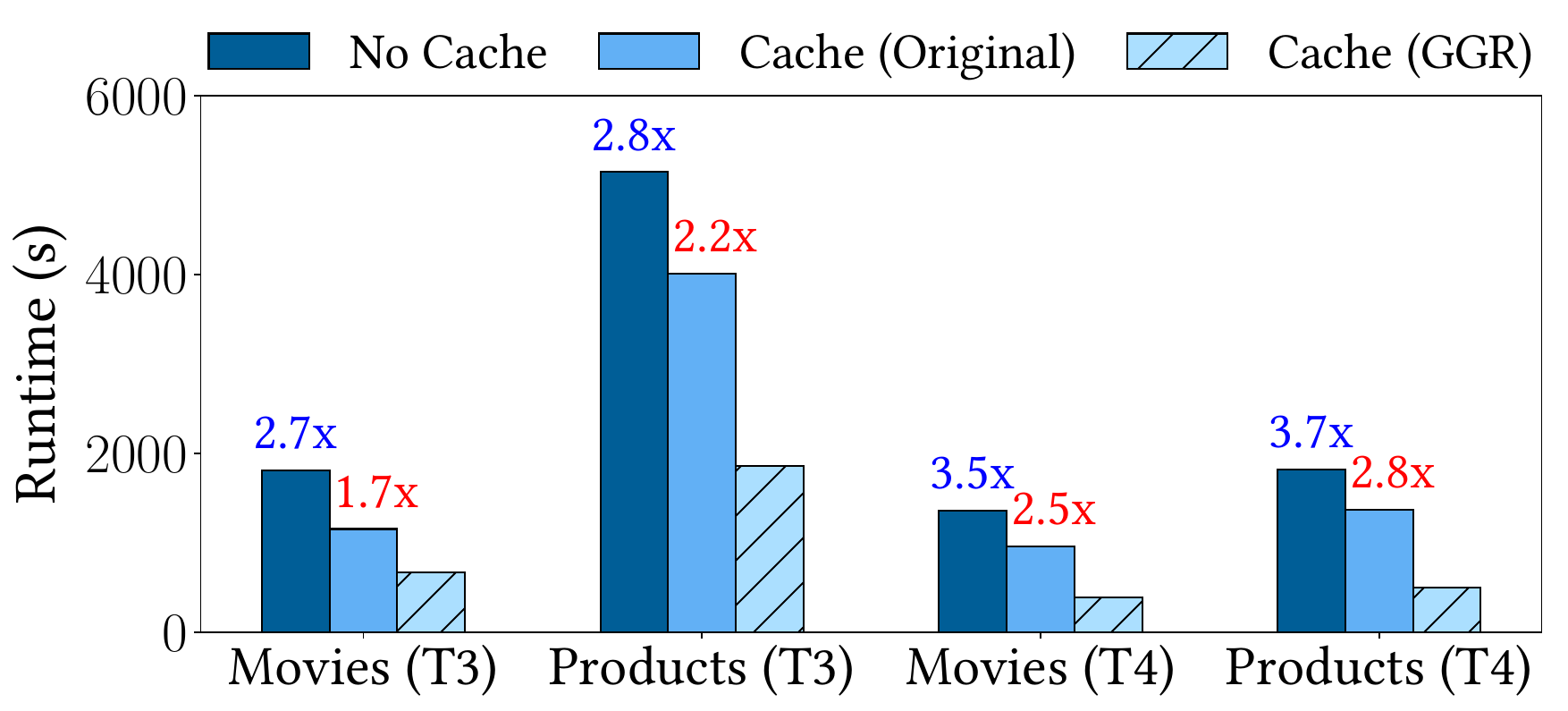}
    \vspace{-1em}
    \caption{End-to-end Result (Multi-LLM Invocation, Aggregation): Our optimizations Cache (\greedy) achieve 1.7 - 2.8$\times$ speed-up over Cache (Original), and 2.7 - 3.7$\times$ speed-up over No Cache. }
    \label{fig:q3q4}

\end{figure}

\textbf{\textit{Overview}}. Fig~\ref{fig:q1q5} and Fig~\ref{fig:q3q4} show the end-to-end latency results of our techniques on LLM filter, projection, multi-LLM invocation, aggregation, and RAG queries with the Llama-3-8B-Instruct model on a single L4. Our evaluation shows that our approach can achieve 1.5 to 3.4$\times$ speedup over Cache (Original) and 1.8 to 3.8$\times$ speedup over No Cache across 16 queries. 
We discuss the evaluation for each query type in detail as below.


\vspace{-0.5em}
\noindent \textbf{\textit{LLM filter.}} 
This query type uses an LLM operator to filter rows, often producing concise outputs of only a few tokens (see Table~\ref{tab:dataset}). Examples include question-answering tasks limited to 'Yes' or 'No' responses, or sentiment labels like 'Positive,' 'Negative,' or 'Neutral.' 
We construct five such queries on the datasets shown in Fig~\ref{fig:filter-q}. 
Our Cache (\greedy) approach achieves a 2.1 -- 3.8$\times$ speed-up over No Cache by caching repeated prefixes from system prompts and input data. 
Cache (Original) with prompt caching enabled can achieve a modest speedup of 1.03 -- 1.9$\times$ over No Cache by reusing instruction prompts and repeated values from the default input table. 
For queries with short decode stages, the primary benefit of prompt caching is the saved prefill computations. 
Our Cache (\greedy) algorithm further reduces end-to-end latency by 1.8 -- 3.0$\times$ over Cache (Original) through reordering rows and fields in the input table to maximize prefix reuse. 

Most review datasets, such as Movies, Products, and BIRD, contain highly distinct values in the first few default fields due to the joining of reviews with metadata tables.
For instance, these tables often begin with a \texttt{review\_content} field. 
Our algorithm prioritizes fields with repeated values, like \texttt{description} and \texttt{product\_title}, leading to a 57 -- 74\% increase in prefix hit rates and a 2.5 -- 3$\times$ speed-up over the original ordering. 
PDMX is a dataset containing 57 fields with many unique, lengthy text entries. In this dataset, our algorithm raises the hit rate from an initial 12\% to 57\%, resulting in a 1.8$\times$ reduction in end-to-end latency. This lower speed-up is due to the nature of long input and 43\% of cache miss from this dataset even for Cache (GGR).
The Beer dataset contains some duplicated values in early fields like \texttt{review/profileName} and Cache (Original) can achieve an initial hit rate of 50\%. Cache (\greedy) can further increase the hit rate by an additional 30\% to reach 80\% and achieve a 2$\times$ speedup.

\vspace{-0.5em}
\noindent \textbf{\textit{LLM projection.}} 
This query type applies the LLM to the selected
data for a specific task, producing longer outputs ranging from 29 to 107 tokens (see Table~\ref{tab:dataset}). 
For example, LLMs can be used to summarize the positive aspects of movies leading to favorable ratings in the Movies dataset.
As shown in Fig~\ref{fig:selection-rag}, for datasets except for SQuAD and FEVER (i.e. RAG queries), Cache (GGR) achieves 2.4$\times$ to 3.7$\times$ speed-up over No Cache, and 1.5$\times$ to 3.4$\times$ speed-up over Cache (Original). 
Notice that as the output token length increases, query execution time across all baselines also grows. 
In cases where the decode stage dominates, benefits from prefill caching are less pronounced, leading to smaller relative performance gains than with LLM Filter queries with shorter output length. 
However, for datasets like BIRD and PDMX, which contain long strings, prompt caching saves memory during the decode stage, making the speedup more noticeable with longer decode times.


\begin{table}[t!]
\footnotesize
\setlength{\tabcolsep}{6pt} 
\begin{tabularx}{\columnwidth}{l@{\hskip 2pt}c@{\hskip 2pt}c@{\hskip 2pt}c@{\hskip 2pt}c@{\hskip 2pt}c@{\hskip 2pt}c@{\hskip 2pt}c}
\toprule
\textbf{Method} & \textbf{Movies} & \textbf{Prods.} & \textbf{BIRD} & \textbf{PDMX} & \textbf{Beer} & \textbf{FEVER} & \textbf{SQuAD} \\
\midrule
\textbf{Original}  & 35\%            & 27\%               & 10\%           & 12\%          & 50\%           & 11\%           & 11\%  \\
\textbf{GGR}    & 86\%             & 83\%               & 85\%           & 57\%           & 80\%           & 67\%           & 70\%  \\
\bottomrule
\end{tabularx}

\caption{PHR (\%) of LLM Filter and RAG queries for Original and GGR, which achieves 30 -- 75\% higher hit rates.}
\label{tab:hit-rate}

\end{table}

\vspace{-0.5em}
\noindent \textbf{\textit{Multi-LLM invocation.}} This query type combines Filter and Selection operations, beginning with an initial LLM filter (e.g., selecting positive reviews), followed by an LLM summarization of the filtered table. 
Applied to the Movies and Products datasets, as shown in Fig~\ref{fig:q3q4}, Cache (\greedy) achieves a 2.7$\times$ and 2.8$\times$ speedup over the No Cache baseline for Movies and Products, respectively. Compared to Cache (Original), Cache (\greedy) attains a speedup of 1.7$\times$ and 2.2$\times$. The relative speedup compared to Cache (Original) reduces for both datasets compared to Filter and Projection queries. This is because the first LLM invocation for filtering is over distinct reviews for sentiment analysis, so Cache (Original) and Cache (\greedy) performance will be similar, reducing the overall benefits. For Movies, this number reduces from 2.5$\times$ to 1.7$\times$ as the first invocation accounts for nearly half the query time; while for Products, the second invocation on Projection dominates runtime due to long decode output length (i.e., around 107), so we can still see 2.2$\times$ speed-up over Cache (Original).

\vspace{-0.5em}
\noindent \textbf{\textit{LLM aggregation.}} This query type uses \texttt{AVG} operator to aggregate the sentiment score on the reviews column with additional columns provided as context. We achieve a 3.5$\times$ speed-up in the Movies dataset and a 3.7$\times$ speed-up in the Products dataset over the No Cache baseline. We also achieve 2.5$\times$ speed-up on Movies and 2.8$\times$ speed-up on Products over Cache (Original). The results of this query type are similar to filtering query results, as the average output length is similar.

\begin{figure}[t!]
    \centering
    \includegraphics[width=0.9\columnwidth]{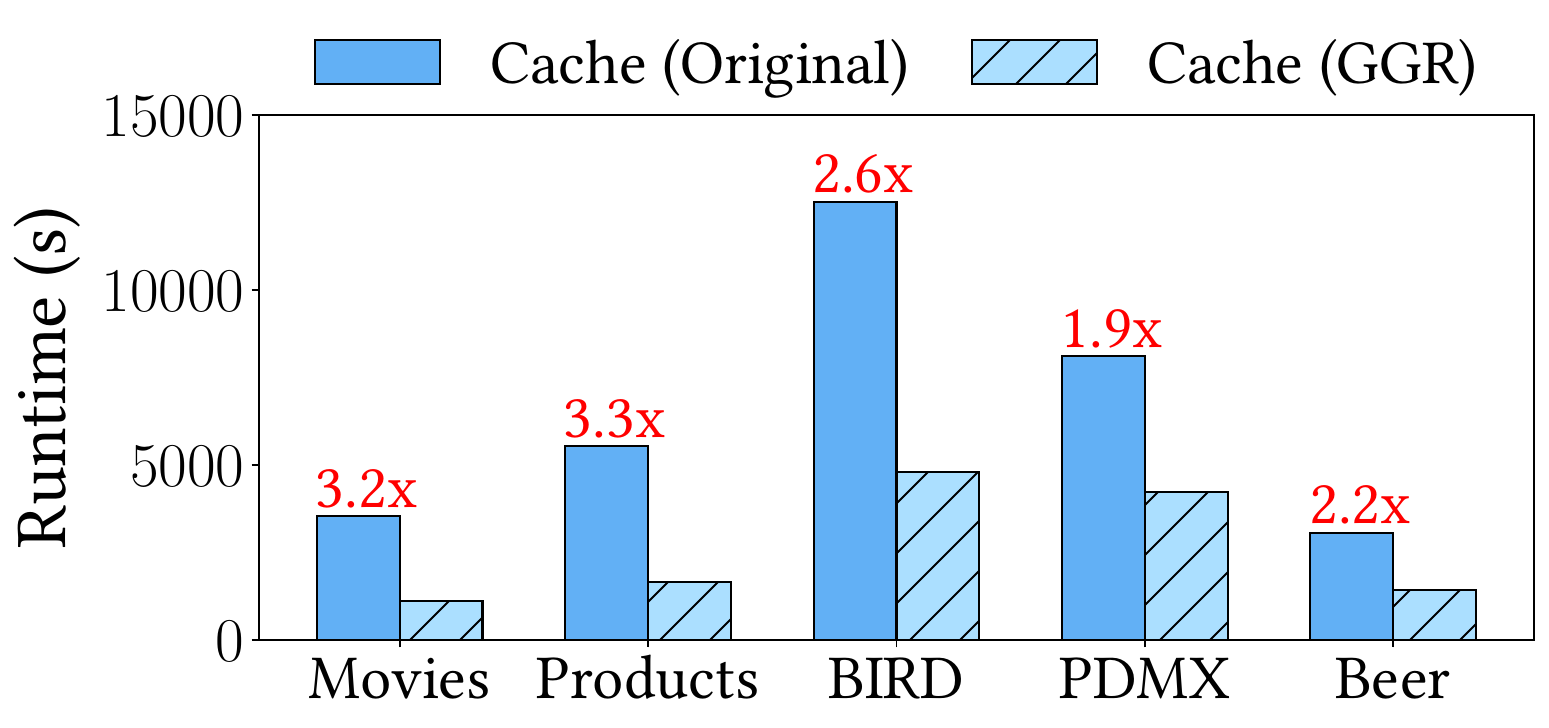}
    \vspace{-1em}
    \caption{Cache (GGR) is able to achieve 1.9 -- 3.3$\times$ speed-up over Cache (Original) for filter queries on Llama3-70B.}
    \label{fig:modelablation}
    \vspace{-2em}
\end{figure}

\vspace{-0.5em}
\noindent \textbf{\textit{RAG.}} This query is performed on a table of questions and the top four to five supporting evidence items extracted from the FEVER and SQuAD datasets. Cache (GGR) achieves a 1.9$\times$ speed-up on both FEVER and SQuAD over the No Cache baseline. We also achieve a 1.8$\times$ speed-up on FEVER and 1.7$\times$ on SQuAD over Cache (Original). 
In this experiment, we embed all supporting contexts for a question/claim into a vector index. We perform a K-nearest neighbor search on the vector index for each question to fetch relevant contexts.
At runtime, we apply our \greedy algorithm to the table of questions and contexts to maximize cache hits. Cache (\greedy) can achieve 56 -- 59\% prefix hit rate improvements over Cache (Original), as multiple questions might share similar contexts, and Cache (\greedy) can rearrange contexts to maximize prefix reuse. 
\vspace{-0.5em}
\textbf{Results on Different Model Sizes} Fig~\ref{fig:modelablation} shows the evaluation of our Cache (\greedy) method compared with Cache (original) on filtering queries, using Llama-3-70B-Instruct with 70B parameters. We run this model on an 8$\times$L4 instance with tensor parallelism and measure the end-to-end query latency. Cache (\greedy) achieves 1.9$\times$ to 3.3$\times$ speed-up under this setup, showing a trend similar compared to the Llama-3-8B model. We evaluate the larger model accuracy on LLM Filter queries in Sec~\ref{sec:accuracy}. We also show results for the smaller 1B model in Appendix~\ref{appendix:models}.

\renewcommand{\arraystretch}{1} 

\begin{figure*}[tbp]
     \centering
     \begin{subfigure}[b]{0.33\textwidth}
        \centering
        \includegraphics[width=\textwidth]{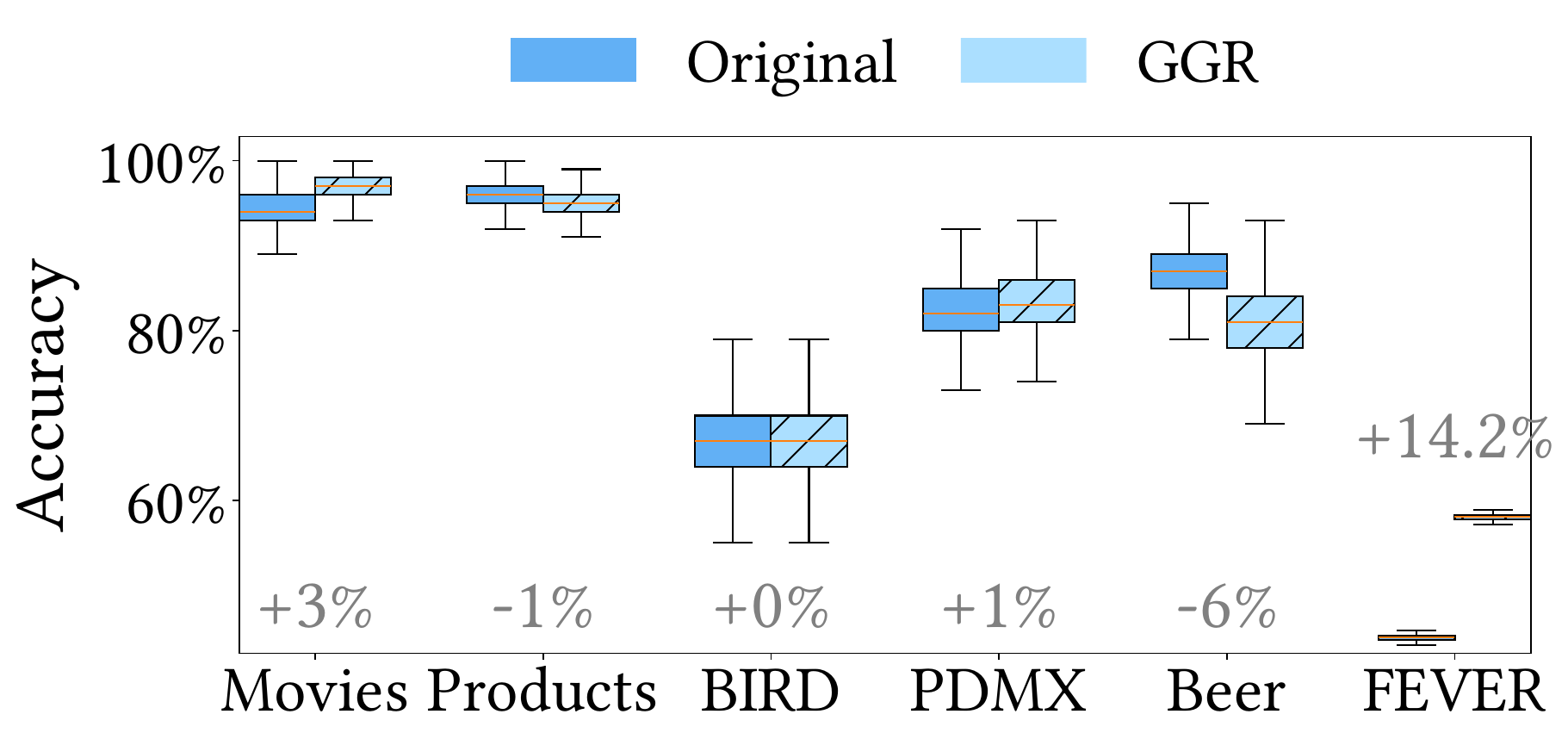}
        \caption{Meta-Llama-3-8B-Instruct}
        \label{fig:movies-runtimes}
    \end{subfigure}
    \hfill
    \begin{subfigure}[b]{0.33\textwidth}
        \centering
        \includegraphics[width=\textwidth]{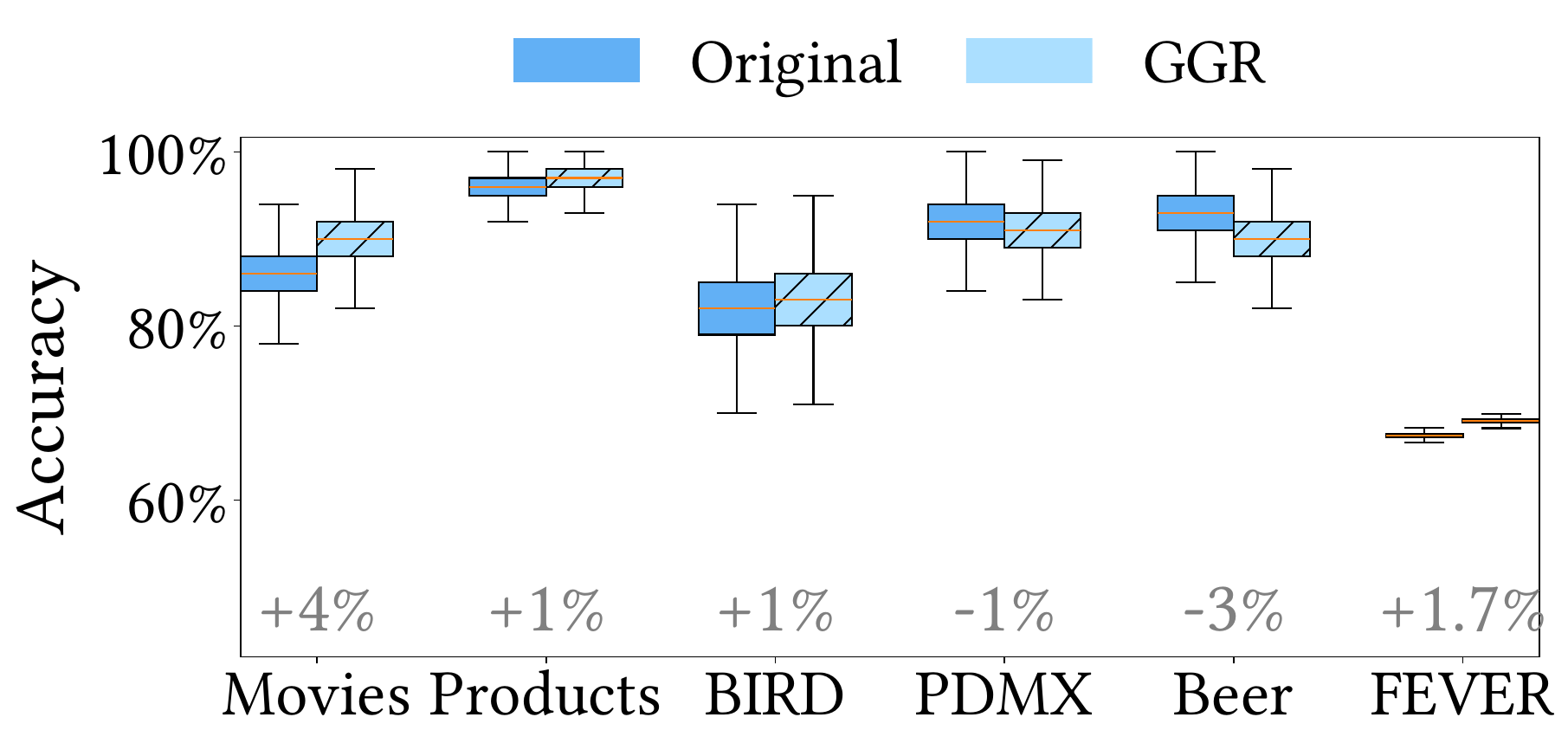}
        \caption{Meta-Llama-3-70B-Instruct}
        \label{fig:products-runtimes}
    \end{subfigure}
    \begin{subfigure}[b]{0.33\textwidth}
        \centering
        \includegraphics[width=\textwidth]{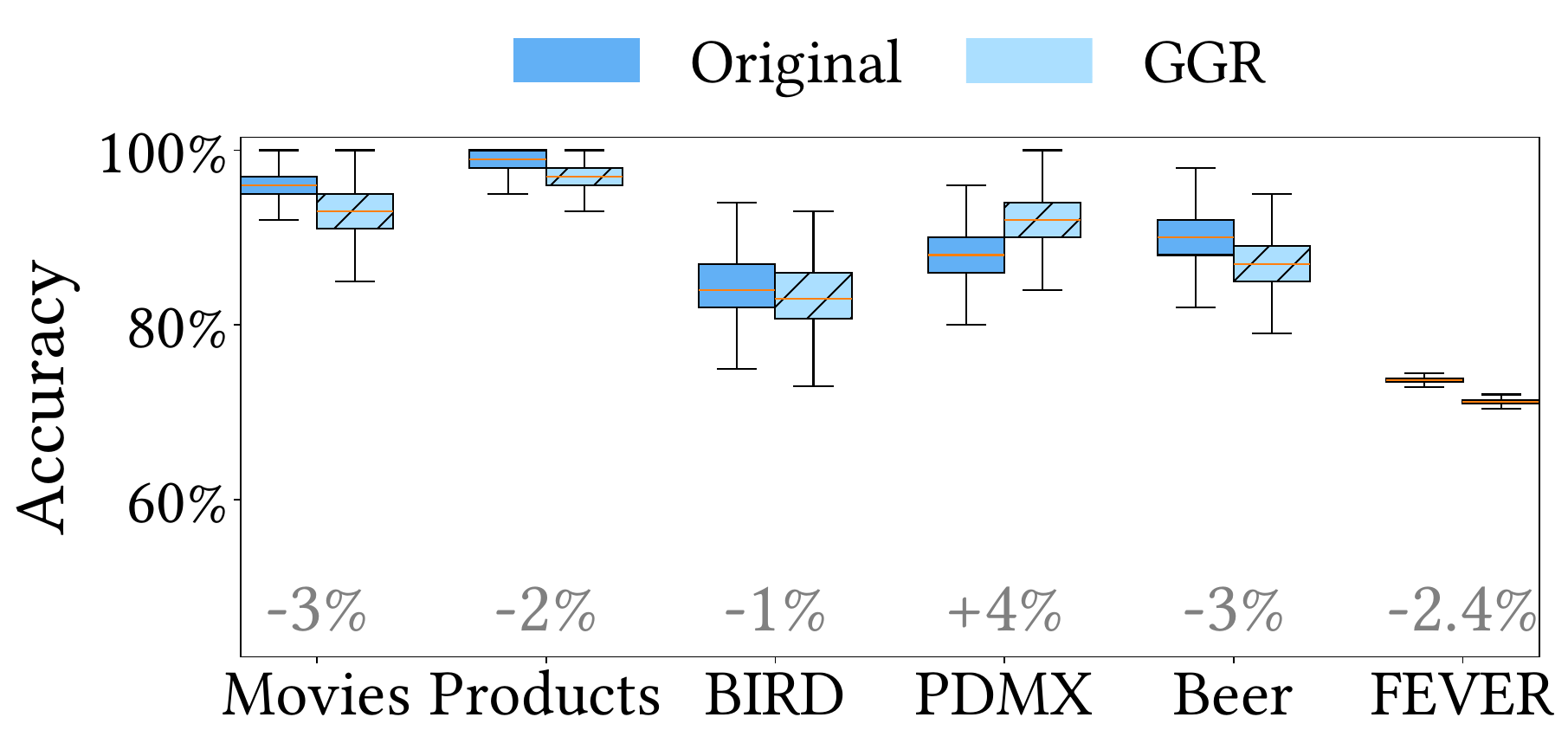}
        \caption{OpenAI GPT-4o}
        \label{fig:products-runtimes}
    \end{subfigure}

    \vspace{-1em}
    \caption{Accuracy of original v.s. \greedy ordering: we perform statistical bootstrapping to get a distribution of exact match accuracy measurements across 10,000 runs. The numbers indicate the difference in the median accuracy of \greedy compared to the original ordering.}
    \label{fig:accuracy}
    \vspace{-0.5em}
\end{figure*}

\vspace{-0.5em}

\subsection{Cost Savings on Proprietary API Endpoints}
This section evaluates the cost efficiency of our \greedy algorithm with closed models that support prompt caching. 
For OpenAI, cached prompts are offered at a 50\% discount compared to uncached prompts.  
Anthropic beta prompt caching~\cite{anthropicpromptcaching} requires users to manually specify prompts to cache. Writing to the cache costs 25\% more than the base input token price for any given model while using cached content costs only 10\% of the base rate. We evaluate OpenAI GPT-4o-mini and Anthropic Claude 3.5 Sonnet, using their pricing models in our cost calculations.\footnote{GPT-4o-mini charges \$0.075/1M tokens for cached tokens versus \$0.15/1M tokens for uncached tokens.}\footnote{Claude 3.5 Sonnet standard input tokens are priced at \$3 per million tokens, cache writes at \$3.75 per million, and cache reads at \$0.30 per million tokens.}




\begin{table}[h!]
\centering
\small
\renewcommand{\arraystretch}{1.2} 

\begin{tabular}{c c c c c c}
\hline
\textbf{Dataset} & \textbf{Model} & \textbf{Method} & \textbf{PHR (\%)} & \textbf{Cost (\$)} & \textbf{Savings (\%)} \\
\hline
\multirow{4}{*}{FEVER} 
& \multirow{2}{*}{4o-mini} & Original & 0.0   & 0.81  & -     \\
&                              & GGR      & 62.2  & 0.55 & 32\%  \\
\cline{2-6}
& \multirow{2}{*}{Sonnet} & Original & 0.0   & 5.49  & -     \\
&                                & GGR      & 30.6  & 4.33  & 21\%  \\
\hline
\end{tabular}
\caption{OpenAI and Anthropic Costs: cache hit rate (HR\%), cost, and savings comparison of GGR over Original for GPT-4o-mini and Claude 3.5 Sonnet in FEVER.}
\label{tab:cost_comparison}

\end{table}

\begin{table}[t!]
\centering
\small
\begin{tabularx}{\columnwidth}{l@{\hskip 20pt}c@{\hskip 20pt}c@{\hskip 8pt}c@{\hskip 20pt}c@{\hskip 8pt}c}
\toprule
\multirow{2}{*}{\textbf{Dataset}} & \multicolumn{2}{c}{\small \textbf{PHR (\%)}} & \multicolumn{2}{c}{\small \textbf{Est. Cost Savings (\%)}} \\
\cmidrule(lr){2-3} \cmidrule(lr){4-5}
                  & \small Original & \small GGR & \small OpenAI & \small Anthropic \\ 
\midrule
\textbf{Movies}  & 34.6          & 85.7         & 31            & 73              \\ 
\textbf{Products}& 26.7          & 83.3         & 33            & 73              \\ 
\textbf{BIRD}    & 10.4          & 84.8         & 39            & 79              \\ 
\textbf{PDMX}    & 11.8          & 56.6         & 24            & 48              \\ 
\textbf{Beer}    & 49.9          & 80.1         & 20            & 55              \\ 
\textbf{FEVER}   & 11.2          & 67.4         & 30            & 60              \\ 
\textbf{SQuAD}   & 11.0          & 69.7         & 31            & 63              \\ 
\bottomrule
\end{tabularx}

\vspace{-0.5em}
\caption{Estimated cost savings: across datasets using PHR from Sec~\ref{sec:end-to-end} and OpenAI and Anthropic's pricing model. }
\label{tab:algoresults}
\vspace{-1.75em}
\end{table}
Since both OpenAI and Anthropic require a minimum prefix length of 1,024 tokens for caching, we duplicate each field value five times, approximating a more realistic dataset with detailed conversations and descriptions.
We select the FEVER dataset for its long input length and use 1000 rows from this dataset. 
For Anthropic experiments, we specify cache write for only the first 1,024 tokens per request as a conservative assumption, as Anthropic does not support automatic prefix detection. 

We evaluate \greedy reordering on two tables submitted to the OpenAI and Anthropic APIs (each row is a request): one reordered with \greedy and one in the original row and field order. 
Table~\ref{tab:cost_comparison} shows that \greedy achieves 32\% cost savings with GPT-4o-mini and 21\% savings with Claude 3.5 Sonnet. 
The hit rate in OpenAI for \greedy-reordered table is 62.2\%, closely matching the hit rate (i.e., 67\%) measured from our previous experiment in Table~\ref{tab:hit-rate}. 
Original ordering receives no cached tokens with 0\% cache hits, as the shared prefix does not meet the 1,024-token minimum.
The Anthropic cache hit rate is around 30.6\%, two times lower than the OpenAI hit rate due to our conservative caching threshold. 

Assume that in the future, automatic prefix caching is enabled and prompts can be cached at arbitrary token lengths. We use the hit rate numbers collected from our previous experiments in Table~\ref{tab:hit-rate} to simulate cost-saving ratios achievable by GGR, compared to the original unordered algorithm. \greedy yields 20 to 39\% cost savings under the OpenAI pricing model and up to 79\% cost savings with Anthropic.

\vspace{-0.5em}
\subsection{Impact of Reordering on Accuracy} \label{sec:accuracy}
As \greedy order alters the input prompt to the LLM, we assess the impact this has on query accuracy using LLM Filter queries (Sec~\ref{subsec:llmqueries}) with constrained output. We also evaluate a RAG query of FEVER, excluding SQuAD due to its open-ended questions. 
FEVER includes ground-truth labels for all records, while 100 rows from other datasets are manually labeled. Using statistical boostrapping~\cite{bootstrapping}, we perform 10K runs, sampling with replacement on each run to obtain a  distribution of accuracy results. Accuracy experiments are conducted with Llama-3-8B-Instruct, Llama-3-70B-Instruct, and GPT-4o models, measured as the percentage of exact matches between the LLM output and the ground truth labels.

In Fig~\ref{fig:accuracy}, we plot the accuracy distributions across the bootstrap runs and the relative difference in median accuracy of \greedy versus original ordering. The accuracy distribution of \greedy ordering is within 5\% accuracy of the original ordering, with the only exception being FEVER with Llama-3-8B, where the ordering with \greedy performs 14.2\% \emph{better} than the original. This is due to the \greedy algorithm places the ``claim'' field at the end of the prompt instead of at the beginning, which Llama3-8B prefers. However, the same behavior does not hold for the larger models. Overall, we can see that larger models like Llama-3-70B and GPT-4o are within 5\% of accuracy difference compared with original ordering and are more robust to field reordering.

\vspace{-0.2em}
\subsection{Algorithm Overhead} 


\begin{table}[t!]
\vspace{1em}
\centering
\footnotesize
\renewcommand{\arraystretch}{1.1} 

\begin{tabular}{c}
\hline
\multicolumn{1}{c}{\textbf{Solver Time (s)}} \\
\begin{tabular}{ccccccc}
Movies & Products & BIRD & PDMX & Beer & FEVER & SQuAD \\
\hline
3.3 & 4.5 & 1.2 & 12.6 & 8.0 & 5.6 & 4.5 \\  
\hline
\end{tabular} \\
\hline
\end{tabular}
\vspace{-0.1em}
\caption{\greedy Solver time (s): \greedy runs under 15 seconds for datasets with up to 30K rows and 57 fields.}
\label{tab:algosolvertimes}
\vspace{-2em}
\end{table}
\textbf{Latency} Table~\ref{tab:algosolvertimes} shows the average overheads of \greedy across datasets, using a row recursion depth of four and column recursion depth of two, or an early stopping threshold of 0.1M hit count. In all cases, \greedy runs in under 15 seconds -- less than 0.01\% of LLM query runtimes. \newline 
\textbf{Memory} \greedy only requires the input table $T$ ($n$ rows, $m$ columns) touched by the query to be loaded into memory. Recursive splitting reduces table size at each step, keeping total memory usage at $O(n \times m)$, aside from minimal stack and temporary variable overhead.

\section{Related Work}

Our optimizations build on recent work in LLM inference as well as prior work integrating machine learning and data management. We describe several major related areas below.

\vspace{-0.5em}
\noindent \textbf{Inference-optimized systems.} There has been a recent rise of dedicated systems for LLM inference, including FasterTransformer \cite{faster-transformers}, Orca \cite{orca-continous-batching}, vLLM \cite{vllm}, and SGLang \cite{sglang}. 
Many systems already explore developing memory-efficient GPU kernels that perform inference while leveraging shared prefixes. 
SGLang's RadixAttention \cite{sglang}, Hydragen \cite{hydragen}, and Cascade Inference \cite{cascade-inference} all implement optimized kernels. 
Our work builds upon prior work investigating high-throughput LLM inference and prefix caching for model serving. In addition, we leverage full workload information from batch queries to further improve performance in relational workloads.

\vspace{-0.5em}
\noindent \textbf{LLMs in Relational Data Analytics} 
Many systems support calling LLMs as operators on relational data, spanning from production database vendors like Databricks \cite{databricks-ai-functions}, Google BigQuery \cite{google-bigquery-llm} and AWS Redshift \cite{aws-redshift-llm} to programming frameworks like LOTUS \cite{lotus}. While these works provide APIs for running LLMs over relational data, they do not explore how reordering data can optimize KV cache hits. 
There is also a line of work ~\cite{noscope, prob-pred} that explores using cheaper models for approximate query generation. This orthogonal direction is not considered in our paper scope, as our work specifically focuses on calling LLMs as functions from inside a regular, given SQL query.



\section{Conclusion}
In this paper, we introduce techniques to optimize LLM invocations in relational data analytics workloads.
By leveraging workload information coupled with observations about the LLM inference process, we can significantly improve end-to-end query performance and reduce costs without affecting query semantics. 
Our technique achieves up to 3.4$\times$ decreases in end-to-end query latency with Llama-3-8B and Llama-3-70B and also achieves up to 32\% cost savings under OpenAI and Anthropic pricing models. 

\section*{Acknowledgement}
We thank Soujanya Ponnapalli for helpful discussions and feedback, and Jelani Nelson for academic advising. This research was supported by gifts from Accenture, AMD, Anyscale, Broadcom Inc., Google, IBM, Intel, Intesa Sanpaolo, Lambda, Mibura Inc, Samsung SDS, and SAP.


\bibliography{reference}

\begin{thebibliography}{41}
\providecommand{\natexlab}[1]{#1}
\providecommand{\url}[1]{\texttt{#1}}
\expandafter\ifx\csname urlstyle\endcsname\relax
  \providecommand{\doi}[1]{doi: #1}\else
  \providecommand{\doi}{doi: \begingroup \urlstyle{rm}\Url}\fi

\bibitem[aws()]{aws-redshift-llm}
{L}arge {L}anguage {M}odels for sentiment analysis with {A}mazon {R}edshift {M}{L} ({P}review) | {A}mazon {W}eb {S}ervices --- aws.amazon.com.
\newblock \url{https://aws.amazon.com/blogs/big-data/large-language-models-for-sentiment-analysis-with-amazon-redshift-ml-preview/}.
\newblock [Accessed 01-03-2024].

\bibitem[dat()]{databricks-ai-functions}
{A}{I} {F}unctions on {D}atabricks --- docs.databricks.com.
\newblock \url{https://docs.databricks.com/en/large-language-models/ai-functions.html}.
\newblock [Accessed 01-03-2024].

\bibitem[goo()]{google-bigquery-llm}
{L}{L}{M} with {V}ertex {A}{I} only using {S}{Q}{L} queries in {B}ig{Q}uery | {G}oogle {C}loud {B}log --- cloud.google.com.
\newblock \url{https://cloud.google.com/blog/products/ai-machine-learning/llm-with-vertex-ai-only-using-sql-queries-in-bigquery}.
\newblock [Accessed 01-03-2024].

\bibitem[ant(2024)]{anthropicpromptcaching}
Prompt caching with claude.
\newblock \url{https://www.anthropic.com/news/prompt-caching}, 2024.

\bibitem[gem(2024)]{gemini}
Context caching.
\newblock \url{https://ai.google.dev/gemini-api/docs/caching?lang=python}, 2024.

\bibitem[lla(2024)]{llama3}
Apr 2024.
\newblock URL \url{https://ai.meta.com/blog/meta-llama-3/}.

\bibitem[Armbrust et~al.(2015)Armbrust, Xin, Lian, Huai, Liu, Bradley, Meng, Kaftan, Franklin, Ghodsi, and Zaharia]{spark-sql}
Armbrust, M., Xin, R.~S., Lian, C., Huai, Y., Liu, D., Bradley, J.~K., Meng, X., Kaftan, T., Franklin, M.~J., Ghodsi, A., and Zaharia, M.
\newblock Spark sql: Relational data processing in spark.
\newblock In \emph{Proceedings of the 2015 ACM SIGMOD International Conference on Management of Data}, SIGMOD '15, pp.\  1383–1394, New York, NY, USA, 2015. Association for Computing Machinery.
\newblock ISBN 9781450327589.
\newblock \doi{10.1145/2723372.2742797}.
\newblock URL \url{https://doi.org/10.1145/2723372.2742797}.

\bibitem[Armbrust et~al.(2020)Armbrust, Das, Sun, Yavuz, Zhu, Murthy, Torres, van Hovell, Ionescu, {\L}uszczak, et~al.]{deltalake}
Armbrust, M., Das, T., Sun, L., Yavuz, B., Zhu, S., Murthy, M., Torres, J., van Hovell, H., Ionescu, A., {\L}uszczak, A., et~al.
\newblock Delta lake: high-performance acid table storage over cloud object stores.
\newblock \emph{Proceedings of the VLDB Endowment}, 13\penalty0 (12):\penalty0 3411--3424, 2020.

\bibitem[Chase(2022)]{langchain}
Chase, H.
\newblock {LangChain}, October 2022.
\newblock URL \url{https://github.com/langchain-ai/langchain}.

\bibitem[Chen et~al.(2012)Chen, Zhang, Liu, Poon, and Wang]{mdc}
Chen, T., Zhang, N.~L., Liu, T., Poon, K.~M., and Wang, Y.
\newblock Model-based multidimensional clustering of categorical data.
\newblock \emph{Artificial Intelligence}, 176\penalty0 (1):\penalty0 2246--2269, 2012.

\bibitem[D{\v{z}}eroski(2003)]{multirelation}
D{\v{z}}eroski, S.
\newblock Multi-relational data mining: an introduction.
\newblock \emph{ACM SIGKDD Explorations Newsletter}, 5\penalty0 (1):\penalty0 1--16, 2003.

\bibitem[Gim et~al.(2024)Gim, Chen, seob Lee, Sarda, Khandelwal, and Zhong]{promptcache}
Gim, I., Chen, G., seob Lee, S., Sarda, N., Khandelwal, A., and Zhong, L.
\newblock Prompt cache: Modular attention reuse for low-latency inference, 2024.
\newblock URL \url{https://arxiv.org/abs/2311.04934}.

\bibitem[He \& McAuley(2016)He and McAuley]{amazon-product-review-dataset}
He, R. and McAuley, J.
\newblock Ups and downs: Modeling the visual evolution of fashion trends with one-class collaborative filtering.
\newblock In \emph{Proceedings of the 25th International Conference on World Wide Web}, WWW '16, pp.\  507–517, Republic and Canton of Geneva, CHE, 2016. International World Wide Web Conferences Steering Committee.
\newblock ISBN 9781450341431.
\newblock \doi{10.1145/2872427.2883037}.
\newblock URL \url{https://doi.org/10.1145/2872427.2883037}.

\bibitem[Huggingface(2023)]{tgi}
Huggingface.
\newblock {Text Generation Inference}, 2023.
\newblock URL \url{https://huggingface.co/docs/text-generation-inference/en/index}.

\bibitem[Idreos et~al.(2007)Idreos, Kersten, Manegold, et~al.]{craking}
Idreos, S., Kersten, M.~L., Manegold, S., et~al.
\newblock Database cracking.
\newblock In \emph{CIDR}, volume~7, pp.\  68--78, 2007.

\bibitem[Ilyas et~al.(2004)Ilyas, Markl, Haas, Brown, and Aboulnaga]{correlation}
Ilyas, I.~F., Markl, V., Haas, P., Brown, P., and Aboulnaga, A.
\newblock Cords: Automatic discovery of correlations and soft functional dependencies.
\newblock In \emph{Proceedings of the 2004 ACM SIGMOD international conference on Management of data}, pp.\  647--658, 2004.

\bibitem[Johnson et~al.(2019)Johnson, Douze, and J{\'e}gou]{johnson2019billion}
Johnson, J., Douze, M., and J{\'e}gou, H.
\newblock Billion-scale similarity search with {GPUs}.
\newblock \emph{IEEE Transactions on Big Data}, 7\penalty0 (3):\penalty0 535--547, 2019.

\bibitem[Juravsky et~al.(2024)Juravsky, Brown, Ehrlich, Fu, Ré, and Mirhoseini]{hydragen}
Juravsky, J., Brown, B., Ehrlich, R., Fu, D.~Y., Ré, C., and Mirhoseini, A.
\newblock Hydragen: High-throughput llm inference with shared prefixes, 2024.

\bibitem[Kang et~al.(2017)Kang, Emmons, Abuzaid, Bailis, and Zaharia]{noscope}
Kang, D., Emmons, J., Abuzaid, F., Bailis, P., and Zaharia, M.
\newblock Noscope: optimizing neural network queries over video at scale.
\newblock \emph{Proc. VLDB Endow.}, 10\penalty0 (11):\penalty0 1586–1597, aug 2017.
\newblock ISSN 2150-8097.

\bibitem[Kwon et~al.(2023)Kwon, Li, Zhuang, Sheng, Zheng, Yu, Gonzalez, Zhang, and Stoica]{vllm}
Kwon, W., Li, Z., Zhuang, S., Sheng, Y., Zheng, L., Yu, C.~H., Gonzalez, J., Zhang, H., and Stoica, I.
\newblock Efficient memory management for large language model serving with pagedattention.
\newblock In \emph{Proceedings of the 29th Symposium on Operating Systems Principles}, SOSP '23, pp.\  611–626, New York, NY, USA, 2023. Association for Computing Machinery.
\newblock ISBN 9798400702297.
\newblock \doi{10.1145/3600006.3613165}.
\newblock URL \url{https://doi.org/10.1145/3600006.3613165}.

\bibitem[Lemire \& Kaser(2011)Lemire and Kaser]{lemire2011reordering}
Lemire, D. and Kaser, O.
\newblock Reordering columns for smaller indexes.
\newblock \emph{Information Sciences}, 181\penalty0 (12):\penalty0 2550--2570, 2011.

\bibitem[Lewis et~al.(2021)Lewis, Perez, Piktus, Petroni, Karpukhin, Goyal, Küttler, Lewis, tau Yih, Rocktäschel, Riedel, and Kiela]{retrieval-augmented-generation}
Lewis, P., Perez, E., Piktus, A., Petroni, F., Karpukhin, V., Goyal, N., Küttler, H., Lewis, M., tau Yih, W., Rocktäschel, T., Riedel, S., and Kiela, D.
\newblock Retrieval-augmented generation for knowledge-intensive nlp tasks, 2021.

\bibitem[Li et~al.(2024)Li, Hui, Qu, Yang, Li, Li, Wang, Qin, Geng, Huo, et~al.]{li2024can}
Li, J., Hui, B., Qu, G., Yang, J., Li, B., Li, B., Wang, B., Qin, B., Geng, R., Huo, N., et~al.
\newblock Can llm already serve as a database interface? a big bench for large-scale database grounded text-to-sqls.
\newblock \emph{Advances in Neural Information Processing Systems}, 36, 2024.

\bibitem[Li et~al.(2023)Li, Zhang, Zhang, Long, Xie, and Zhang]{li2023towards}
Li, Z., Zhang, X., Zhang, Y., Long, D., Xie, P., and Zhang, M.
\newblock Towards general text embeddings with multi-stage contrastive learning.
\newblock \emph{arXiv preprint arXiv:2308.03281}, 2023.

\bibitem[Long et~al.(2024)Long, Novack, Berg-Kirkpatrick, and McAuley]{pdmx}
Long, P., Novack, Z., Berg-Kirkpatrick, T., and McAuley, J.
\newblock Pdmx: A large-scale public domain musicxml dataset for symbolic music processing, 2024.
\newblock URL \url{https://arxiv.org/abs/2409.10831}.

\bibitem[Lu et~al.(2018)Lu, Chowdhery, Kandula, and Chaudhuri]{prob-pred}
Lu, Y., Chowdhery, A., Kandula, S., and Chaudhuri, S.
\newblock Accelerating machine learning inference with probabilistic predicates.
\newblock In \emph{Proceedings of the 2018 International Conference on Management of Data}, SIGMOD '18, pp.\  1493–1508, New York, NY, USA, 2018. Association for Computing Machinery.
\newblock ISBN 9781450347037.

\bibitem[McAuley et~al.(2012)McAuley, Leskovec, and Jurafsky]{ratebeer}
McAuley, J., Leskovec, J., and Jurafsky, D.
\newblock Learning attitudes and attributes from multi-aspect reviews, 2012.
\newblock URL \url{https://arxiv.org/abs/1210.3926}.

\bibitem[NVIDIA(2023{\natexlab{a}})]{faster-transformers}
NVIDIA.
\newblock {Faster Transformer}, 2023{\natexlab{a}}.
\newblock URL \url{https://github.com/NVIDIA/FasterTransformer}.

\bibitem[NVIDIA(2023{\natexlab{b}})]{trt-llm}
NVIDIA.
\newblock {TensorRT LLM}, 2023{\natexlab{b}}.
\newblock URL \url{https://github.com/NVIDIA/TensorRT-LLM}.

\bibitem[OpenAI()]{openai-pricing}
OpenAI.
\newblock {P}ricing --- openai.com.
\newblock \url{https://openai.com/pricing}.
\newblock [Accessed 01-03-2024].

\bibitem[Pang \& Lee(2005)Pang and Lee]{rotten-tomatoes-movies-dataset}
Pang, B. and Lee, L.
\newblock Seeing stars: Exploiting class relationships for sentiment categorization with respect to rating scales.
\newblock In \emph{Proceedings of the ACL}, 2005.

\bibitem[Patel et~al.(2024)Patel, Jha, Guestrin, and Zaharia]{lotus}
Patel, L., Jha, S., Guestrin, C., and Zaharia, M.
\newblock Lotus: Enabling semantic queries with llms over tables of unstructured and structured data, 2024.
\newblock URL \url{https://arxiv.org/abs/2407.11418}.

\bibitem[Rajpurkar et~al.(2016)Rajpurkar, Zhang, Lopyrev, and Liang]{squad-dataset}
Rajpurkar, P., Zhang, J., Lopyrev, K., and Liang, P.
\newblock Squad: 100,000+ questions for machine comprehension of text, 2016.

\bibitem[Stonebraker et~al.(2018)Stonebraker, Abadi, Batkin, Chen, Cherniack, Ferreira, Lau, Lin, Madden, O'Neil, et~al.]{stonebraker2018c}
Stonebraker, M., Abadi, D.~J., Batkin, A., Chen, X., Cherniack, M., Ferreira, M., Lau, E., Lin, A., Madden, S., O'Neil, E., et~al.
\newblock C-store: a column-oriented dbms.
\newblock In \emph{Making Databases Work: the Pragmatic Wisdom of Michael Stonebraker}, pp.\  491--518. 2018.

\bibitem[Thorne et~al.(2018)Thorne, Vlachos, Christodoulopoulos, and Mittal]{fever}
Thorne, J., Vlachos, A., Christodoulopoulos, C., and Mittal, A.
\newblock Fever: a large-scale dataset for fact extraction and verification, 2018.
\newblock URL \url{https://arxiv.org/abs/1803.05355}.

\bibitem[Vaswani et~al.(2023)Vaswani, Shazeer, Parmar, Uszkoreit, Jones, Gomez, Kaiser, and Polosukhin]{attention-is-all-you-need}
Vaswani, A., Shazeer, N., Parmar, N., Uszkoreit, J., Jones, L., Gomez, A.~N., Kaiser, L., and Polosukhin, I.
\newblock Attention is all you need, 2023.

\bibitem[Wilcox(2003)]{bootstrapping}
Wilcox, R.~R.
\newblock Bootstrap confidence interval.
\newblock In \emph{Applying Contemporary Statistical Techniques}. Academic Press, 2003.

\bibitem[Ye et~al.(2024)Ye, Lai, Lu, Lin, Zheng, Chen, Chen, and Ceze]{cascade-inference}
Ye, Z., Lai, R., Lu, B.-R., Lin, C.-Y., Zheng, S., Chen, L., Chen, T., and Ceze, L.
\newblock Cascade inference: Memory bandwidth efficient shared prefix batch decoding, February 2024.
\newblock URL \url{https://flashinfer.ai/2024/02/02/cascade-inference.html}.

\bibitem[Yu et~al.(2022)Yu, Jeong, Kim, Kim, and Chun]{orca-continous-batching}
Yu, G.-I., Jeong, J.~S., Kim, G.-W., Kim, S., and Chun, B.-G.
\newblock Orca: A distributed serving system for {Transformer-Based} generative models.
\newblock In \emph{16th USENIX Symposium on Operating Systems Design and Implementation (OSDI 22)}, pp.\  521--538, Carlsbad, CA, July 2022. USENIX Association.
\newblock ISBN 978-1-939133-28-1.
\newblock URL \url{https://www.usenix.org/conference/osdi22/presentation/yu}.

\bibitem[Zaharia et~al.(2012)Zaharia, Chowdhury, Das, Dave, Ma, McCauly, Franklin, Shenker, and Stoica]{zaharia2012resilient}
Zaharia, M., Chowdhury, M., Das, T., Dave, A., Ma, J., McCauly, M., Franklin, M.~J., Shenker, S., and Stoica, I.
\newblock Resilient distributed datasets: A $\{$Fault-Tolerant$\}$ abstraction for $\{$In-Memory$\}$ cluster computing.
\newblock In \emph{9th USENIX symposium on networked systems design and implementation (NSDI 12)}, pp.\  15--28, 2012.

\bibitem[Zheng et~al.(2023)Zheng, Yin, Xie, Huang, Sun, Yu, Cao, Kozyrakis, Stoica, Gonzalez, Barrett, and Sheng]{sglang}
Zheng, L., Yin, L., Xie, Z., Huang, J., Sun, C., Yu, C.~H., Cao, S., Kozyrakis, C., Stoica, I., Gonzalez, J.~E., Barrett, C., and Sheng, Y.
\newblock Efficiently programming large language models using sglang, 2023.

\end{thebibliography}
\bibliographystyle{mlsys2025}
\appendix

\section{Query Examples}
\label{appendix:queries}
Our benchmark suite incorporates a broad range of query types. We show examples of each query type as follows.

\vspace{-0.2em}

\textbf{\textit{LLM filter.}} This query type leverages LLM for filtering data within a \texttt{WHERE} clause. The LLM processes and analyzes information to meet some specified criteria, such as identifying whether a movie is suitable for kids. This query type illustrates typical use cases in sentiment analysis and content filtering, which are important for application tasks, such as customer feedback analysis and content moderation. 

\begin{mdframed}[linecolor=black, linewidth=.5pt]
\begin{minted}[fontsize=\small]{sql}
SELECT t.movietitle
FROM MOVIES
WHERE LLM(
    'Given the following fields, determine whether the movie is suitable for kids. Answer ONLY with "Yes" or "No".',
    movieinfo,
    reviewcontent,
    reviewtype,
    movietitle
) = 'Yes'
\end{minted}
\end{mdframed} 
\vspace{8pt}
\vspace{8pt}
\vspace{-2em}
\textbf{\textit{LLM projection.}} This query type makes calls to an LLM within a \texttt{SELECT} statement to process information from specified database column(s). It reflects common tasks in data analytics in which the LLM is used for summarization and interpretation based on certain data attributes.

\begin{mdframed}[linecolor=black, linewidth=.5pt]
\begin{minted}[fontsize=\small]{sql}
SELECT LLM(
    'Given the following information, summarize good qualities in this movie that led to a favorable rating.',
    reviewcontent, movieinfo
)
FROM MOVIES
\end{minted}
\end{mdframed} 


\textbf{\textit{Multi-LLM invocation.}} This query type involves multiple LLM calls in different parts of the query and addresses scenarios in which several layers of data processing or analysis are required. It represents advanced analytical tasks, such as combining different data insights.

\begin{mdframed}[linecolor=black, linewidth=.5pt]
\begin{minted}[fontsize=\small]{sql}
SELECT LLM(
    'Given the information about a movie, summarize the good qualities that led to a favorable rating.',
    reviewtype,
    reviewcontent,
    movieinfo,
    genres
)
FROM MOVIES
WHERE LLM(
    'Given the following review, answer whether the sentiment is "POSITIVE" or "NEGATIVE". Respond ONLY with "POSITIVE" or "NEGATIVE", in all caps.',
    reviewcontent
) = 'NEGATIVE'
\end{minted}
\end{mdframed} 


\vspace{-1em}
\textbf{\textit{LLM aggregation.}} This query type incorporates an AVG operator that incorporates LLM outputs into further query processing. For example, one
could use LLMs to assign sentiment scores to individual reviews and then aggregate these scores to calculate an average sentiment for overall customer feedback.
This query type is essential for tasks that need to extract insights from complex textual data.

\begin{mdframed}[linecolor=black, linewidth=.5pt]
\begin{minted}[fontsize=\small]{sql}
SELECT AVG(
    LLM(
        'Rate sentiment in numerical values from 1 (bad) to 5 (good).',
        reviewcontent, movieinfo
    )
) AS AverageScore
FROM MOVIES
\end{minted}
\end{mdframed} 
\vspace{-1em}

\textbf{\textit{Retrieval-augmented generation (RAG)}.} This query type leverages external knowledge bases for enhanced LLM processing, enriching LLM queries with a broader context. It simulates use cases where queries need to pull in relevant information from external sources, such as document databases or knowledge graphs, to provide comprehensive answers. 

\begin{mdframed}[linecolor=black, linewidth=.5pt]
\begin{minted}[fontsize=\small]{sql}
SELECT LLM(
    'Given a question and four supporting contexts, answer the provided question.', VectorDB.search(question, k=4), question)
FROM FEVER
\end{minted}
\end{mdframed}

\section{Dataset Information}
\label{appendix:fields}
We detail the fields and functional dependencies 
(FDs) used for each dataset as follows. 
\begin{tcolorbox}[colback=gray!5!white, colframe=black!75!white, title=MOVIES]
\footnotesize
\begin{verbatim}
columns:
genres, movieinfo, movietitle, 
productioncompany, reviewcontent, 
reviewtype, rottentomatoeslink, 
topcritic

FDs: 
movieinfo, movietitle, 
rottentomatoeslink
\end{verbatim}
\end{tcolorbox}

\begin{tcolorbox}[colback=gray!5!white, colframe=black!75!white, title=PRODUCTS]
\footnotesize
\begin{verbatim}
columns: 
description, id, parent_asin, 
product_title, rating, review_title, 
text, verified_purchase


FDs: 
parent_asin, product_title
\end{verbatim}
\end{tcolorbox}

\begin{tcolorbox}[colback=gray!5!white, colframe=black!75!white, title=BIRD]
\footnotesize
\begin{verbatim}
columns:
Body, PostDate, PostId, Text

FDs: 
Body, PostId
\end{verbatim}
\end{tcolorbox}

\begin{tcolorbox}[colback=gray!5!white, colframe=black!75!white, title=PDMX]
\footnotesize
\begin{verbatim}
columns: 
artistname, bestarrangement, bestpath, 
bestuniquearrangement, composername,
complexity, genre, grooveconsistency, 
groups, hasannotations, hascustomaudio,
hascustomvideo, haslyrics, hasmetadata, 
haspaywall, id, isbestarrangement, 
isbestpath, isbestuniquearrangement, 
isdraft, isofficial, isoriginal, 
isuserpro, isuserpublisher, isuserstaff, 
license, licenseurl, metadata, 
nannotations, ncomments, nfavorites, 
nlyrics, notesperbar, nnotes, nratings, 
ntracks, ntokens, nviews, path, 
pitchclassentropy, postdate, postid, 
publisher, rating, scaleconsistency, 
songlength, songlengthbars, 
songlengthbeats, songlengthseconds, 
songname, subsetall, subsetdeduplicated, 
subsetrated, subsetrateddeduplicated, 
subtitle, tags, text, title, tracks, 
version


FDs:
[metadata, path], 
[hasannotations, hasmetadata, isdraft, 
isofficial, isuserpublisher, subsetall
]

\end{verbatim}
\end{tcolorbox}

\begin{tcolorbox}[colback=gray!5!white, colframe=black!75!white, title=BEER]
\footnotesize
\begin{verbatim}
columns: 
beer/beerId, beer/name, beer/style, 
review/appearance, review/overall, 
review/palate, review/profileName, 
review/taste, review/time

FDs: 
[beer/beerId, beer/name]
\end{verbatim}
\end{tcolorbox}

\begin{tcolorbox}[colback=gray!5!white, colframe=black!75!white, title=FEVER]
\footnotesize
\begin{verbatim}
-- FEVER --
columns: 
claim, evidence1, evidence2, 
evidence3, evidence4

FDs: []
\end{verbatim}
\end{tcolorbox}

\begin{tcolorbox}[colback=gray!5!white, colframe=black!75!white, title=SQuAD]
\footnotesize
\begin{verbatim}
columns: 
question, context1, context2,
context3, context4, context5

FDs: []
\end{verbatim}
\end{tcolorbox}

\vspace{-0.5em}
\section{Prompts}
\label{appendix:prompts}
We detail the system and user prompts for each query type and dataset as follows. 
\vspace{-0.5em}


\begin{tcolorbox}[colback=gray!5!white, colframe=black!75!white, title=System Prompt]
\scriptsize
\begin{verbatim}
You are a data analyst. Use the provided JSON data 
to answer the user query based on the specified 
fields. Respond with only the answer, 
no extra formatting. 

Answer the below query: 
{QUERY} 

Given the following data: 
{fields}
\end{verbatim}
\end{tcolorbox}


%
\vspace{-0.6em}


\begin{tcolorbox}[colback=gray!5!white, colframe=black!75!white, title=User Prompt - LLM Aggregation]
\scriptsize
\begin{verbatim}
MOVIES: Given the following fields of a movie 
description and a user review, assign a sentiment 
score for the review out of 5. Answer with ONLY a 
single integer between 1 (bad) and 5 (good).

PRODUCTS: Given the following fields of a product 
description and a user review, assign a sentiment
score for the review out of 5. Answer with ONLY a
single integer between 1 (bad) and 5 (good).
\end{verbatim}
\end{tcolorbox}
\vspace{-0.6em}

\begin{tcolorbox}[colback=gray!5!white, colframe=black!75!white, title=User Prompt - Multi-LLM Invocation]
\scriptsize
\begin{verbatim}
MOVIES/PRODUCTS: Given the following review, answer 
whether the sentiment associated is 'POSITIVE' or 
'NEGATIVE'. Answer in all caps with ONLY 'POSITIVE' 
or 'NEGATIVE': 
\end{verbatim}
\end{tcolorbox}
\vspace{-0.6em}

\begin{tcolorbox}[colback=gray!5!white, colframe=black!75!white, title=User Prompt - LLM Filter]
\scriptsize
\begin{verbatim}
MOVIES: Given the following fields, answer in one 
word, 'Yes' or 'No', whether the movie would be 
suitable for kids.  Answer with ONLY 'Yes' or 'No'.

PRODUCTS: Given the following fields determine if 
the review speaks positively ('POSITIVE'), 
negatively ('NEGATIVE'), or netural ('NEUTRAL') 
about the product. Answer only 'POSITIVE', 
'NEGATIVE', or 'NEUTRAL', nothing else.

BIRD: Given the following fields related to posts 
in an online codebase community, answer whether the
post is related to statistics. Answer with only 
'YES' or 'NO'.

PDMX: Based on following fields, answer 'YES' or 
'NO' if any of the song information references a 
specific individual. Answer only 'YES' or 'NO', 
nothing else.

BEER: Based on the beer descriptions, does this 
beer have European origin? Answer 'YES' if it does 
or 'NO' if it doesn't.
\end{verbatim}
\end{tcolorbox}

\vspace{-0.5em}
\begin{tcolorbox}[colback=gray!5!white, colframe=black!75!white, title=User Prompt - LLM Projection]
\scriptsize
\begin{verbatim}
MOVIES: Given information including movie 
descriptions and critic reviews, summarize the good
qualities in this movie that led to a favorable 
rating. (also used in multi-invocation)

PRODUCTS: Given the following fields related to 
amazon products, summarize the product, then answer 
whether the product description is consistent with 
the quality expressed in the review. (also used 
in multi-invocation)

BIRD: Given the following fields related to posts 
in an online codebase community, summarize how the
comment Text related to the post body.

PDMX: Given the following fields, provide an 
overview on the music type, and analyze the given 
scores. Give exactly 50 words of summary.

BEER: Given the following fields, provide an 
high-level overview on the beer and review in a 
20 words paragraph.
\end{verbatim}
\end{tcolorbox}
\vspace{-0.6em}

\begin{tcolorbox}[colback=gray!5!white, colframe=black!75!white, title=User Prompt - RAG]
\scriptsize
\begin{verbatim}
FEVER: You are given 4 pieces of evidence as 
{evidence1}, {evidence2}, {evidence3}, and 
{evidence4}. You are also given a claim as {claim}. 
Answer SUPPORTS if the pieces of evidence support 
the given {claim}, REFUTES if the evidence refutes 
the given {claim}, or NOT ENOUGH INFO if there is
not enough information to answer. Your answer
should just be SUPPORTS, REFUTES, or NOT ENOUGH
INFO and nothing else.

SQuAD: Given a question and supporting contexts, 
answer the provided question.
\end{verbatim}
\end{tcolorbox}

\section{Ablations}
We present two sets of ablation experiments: one comparing the prefix hit rate (PHR) between \greedy and an optimal oracle, and another examining the impact of using a smaller LLM model.

\subsection{PHR of GGR v.s. \optimal}
\label{appendix:hit-rate}
OPHR is a very expensive brute-force oracle algorithm that iterates through all possible combinations of value groups and calculates the prefix hit count. In our empirical evaluation, it is impractical to run on larger datasets.

Thus, we test the first (10, 25, 50, 100, 200) rows for each dataset and terminate OPHR runs exceeding 2 hours, reporting the result of the successful run with the most rows. For PDMX, we reduce 57 columns to 10 to enable runs on even 
as few as 10 rows. The PHR (prefix hit rate) and solver runtime in seconds across datasets are reported in Table~\ref{tab:phr_runtime_combined}, with the dataset labeled as \textit{\{dataset\}}-\textit{\{\#rows\}}.


\begin{table}[h]
\centering
\footnotesize
\begin{tabular}{lcccccc}
\toprule
\textbf{Dataset} & \multicolumn{3}{c}{\textbf{PHR (\%)}} & \multicolumn{2}{c}{\textbf{Solver Runtime (s)}} \\
\cmidrule(lr){2-4} \cmidrule(lr){5-6}
& \textbf{OPHR} & \textbf{GGR} & \textbf{Diff} & \textbf{OPHR} & \textbf{GGR} \\
\midrule
Movies-50   & 80.6 & 80.6 & 0\%   & 2556  & 0.05 \\
Products-25 & 19.7 & 18.5 & -1.2\% & 357   & 0.06 \\
BIRD-50     & 77.5 & 76.2 & -1.3\% & 0.43  & 0.05 \\
PDMX-25     & 29.4 & 28.6 & -0.8\% & 822   & 0.05 \\
Fever-50    & 7.3  & 6.9  & -0.4\% & 110   & 0.23 \\
Beer-10     & 25.7 & 25.6 & -0.1\% & 1269  & 0.08 \\
SQuAD-10    & 34.0 & 34.0 & 0\%   & 1.6   & 0.05 \\
\bottomrule
\end{tabular}
\caption{Comparison of Prefix Hit Rate (PHR) and solver runtime across datasets. GGR achieves near-optimal PHR while being orders of magnitude faster than OPHR.}
\label{tab:phr_runtime_combined}
\end{table}



We can see that on these small samples of the datasets, our algorithm (GGR) achieves within 2\% of the optimal, but can be up to \textit{hours faster} on solver runtime. 

\subsection{Results of Smaller Model}
\label{appendix:models}
To analyze the impact of using a smaller model, we run the Filter Query described in Fig.~\ref{fig:filter-q} with the Llama-3.2-1B model, using the same setup as with Llama-3 8B (i.e., single L4 instance), and compare the prefix hit rate and end-to-end query execution time of GGR with the default vLLM baseline (i.e. Cache Original). The results are reported in Table~\ref{tab:llama32results}. 


\begin{table}[h]
\centering
\small
\setlength{\tabcolsep}{4pt}
\begin{tabular}{lccc}
\toprule
\textbf{Metric} & \textbf{BIRD} & \textbf{Movies} & \textbf{PDMX} \\
\midrule
Runtime (orig/GGR) & 1.5$\times$ & 1.3$\times$ & 1.3$\times$ \\
Orig PHR (\%) & 10.41 & 29.32 & 11.97 \\
GGR PHR (\%) & 83.99 & 82.10 & 56.00 \\
\midrule
\textbf{Metric} & \textbf{Products} & \textbf{BEER} & \\
\midrule
Runtime (orig/GGR) & 1.4$\times$ & 1.2$\times$ & \\
Orig PHR (\%) & 24.06 & 47.98 & \\
GGR PHR (\%) & 82.10 & 73.93 & \\
\bottomrule
\end{tabular}
\caption{Cache runtime ratio and prefix hit rate (PHR) (\%) comparison between original and GGR ordering for Llama-3.2-1B.}
\label{tab:llama32results}
\end{table}


We observe similar prefix hit rates with Llama-3.2-1B compared to our previous 8B model runs. This consistency arises from the effectiveness of GGR field reordering, which converts non-reusable field contents (0 hits) into reusable prefixes within the cache.
We also observe that under the same GPU instance setup (e.g., L4 with 24 GB memory), larger models like Llama-8B (7.6 GB) exhibit larger relative performance gains from GGR compared to smaller models like Llama-1B (1.8 GB), despite seeing similar prefix hit rates. This is because prefix caching benefits from reducing computational overhead on shared prefixes and enabling larger batch sizes for LLM generation by reducing memory usage through sharing. For smaller models, the availability of ample GPU memory diminishes the relative impact of prefix caching, as larger batch sizes can be achieved without relying on caching. But for larger models, or when there is less available GPU space, prefix caching benefits become more pronounced.


\end{document}